\documentclass{article}

\usepackage[utf8]{inputenc}
\usepackage{amsmath}
\usepackage{makecell}
\usepackage{authblk}

\usepackage{neurips_2023}
\usepackage{algorithm}
\usepackage{algpseudocode}
\usepackage{float}
\usepackage{multirow}
\usepackage{subcaption}
\setcitestyle{numbers,open={[},close={]},aysep={},citesep={,}}
\usepackage{adjustbox}

\usepackage{listings}
\usepackage{xcolor}

\definecolor{codegreen}{rgb}{0,0.6,0}
\definecolor{codegray}{rgb}{0.5,0.5,0.5}
\definecolor{codepurple}{rgb}{0.58,0,0.82}
\definecolor{backcolour}{rgb}{0.95,0.95,0.92}

\lstdefinestyle{mystyle}{
    backgroundcolor=\color{backcolour},   
    commentstyle=\color{codegreen},
    keywordstyle=\color{magenta},
    numberstyle=\tiny\color{codegray},
    stringstyle=\color{codepurple},
    basicstyle=\ttfamily\footnotesize,
    breakatwhitespace=false,         
    breaklines=true,                 
    captionpos=b,                    
    keepspaces=true,                 
    numbers=none,                    
    numbersep=5pt,                  
    showspaces=false,                
    showstringspaces=false,
    showtabs=false,                  
    tabsize=2
}

\usepackage[utf8]{inputenc} 
\usepackage[T1]{fontenc}    
\usepackage[hidelinks]{hyperref}       
\usepackage{url}            
\usepackage{booktabs}       
\usepackage{amsfonts}       
\usepackage{nicefrac} 
\usepackage{authblk} 
\usepackage{microtype}      
\usepackage{xcolor}         
\usepackage{graphicx}
\usepackage{adjustbox}
\usepackage{placeins}
\usepackage{float}
\usepackage{multirow}
\usepackage{tabularx}
\usepackage{authblk}
\usepackage{hyperref}
\usepackage{etoolbox}
\usepackage{array}
\usepackage[bottom]{footmisc}

\title{WebWISE: Web Interface Control and Sequential Exploration with Large Language Models}
\author{%
  Heyi Tao$^{1}$ \thanks{Equal Contribution} \quad Sethuraman T V$^{1*}$ \quad Michal Shlapentokh-Rothman$^1$ \quad Derek Hoiem$^1$\\
  $^1$University of Illinois at Urbana Champaign\\
  \texttt{\{heyitao2,st34,michal5,dhoiem\}@illinois.edu} \\
}

\begin{document}
\maketitle




%
%

\begin{abstract}
The paper investigates using a Large Language Model (LLM) to automatically perform web software tasks using click, scroll, and text input operations. Previous approaches, such as reinforcement learning (RL) or imitation learning, are inefficient to train and task-specific. Our method uses filtered Document Object Model (DOM) elements as observations and performs tasks step-by-step, sequentially generating small programs based on the current observations. We use in-context learning, either benefiting from a single manually provided example, or an automatically generated example based on a successful zero-shot trial. We evaluate the proposed method on the MiniWob++ benchmark. With only one in-context example, our WebWISE method achieves similar or better performance than other methods that require many demonstrations or trials.
\end{abstract}

\section{Introduction}
A major goal of AI is to develop intelligent agents that interact with their environments to perform tasks. This goal is often explored in the context of physical environments. Our work explores performing software tasks with the long-term aim of creating agents that work using software designed for humans. Software tasks are valuable by themselves --- much of the work we do is on computers --- and also offer highly controllable and repeatable tasks that have many of the same challenges as physical tasks, such as manipulable environments, goals that require long sequences of actions, and need for exploration. Prior works to control software have used reinforcement learning, requiring many demonstrations and scored trials to learn simple interaction tasks. Instead, we use a Large Language Model (LLM) to generate actions (e.g., click and enter text) based on environment observations (DOM elements) in web software.

The goal is to complete a web software task, given a natural language instruction and an API for observing and interacting with the environment. In this setting, our observations are DOM elements, indicating the layout and state of buttons, text, and other displayed elements. Actions include clicking an element, scrolling the mouse wheel, and entering text in a text box. The experiments are performed on the MiniWoB++ benchmark~\citep{zheran2018reinforcement}, which consists of randomized simple tasks that involve menu navigation, text entry, and/or clicking buttons and other interactive elements.

LLMs are commonly used to generate code, e.g. Codex~\cite{chen2021evaluating_llms_trained_on_code_codex}, and their use has recently been explored to complete AI tasks given an API and a small number of in-context examples, e.g., VisProg~\citep{Gupta_2023_CVPR}, ViperGPT~\citep{suris2023vipergpt}, and ProgPrompt~\citep{progprompt}. Our application of controlling software differs in two key ways:
\begin{itemize}
\item \textbf{Environment Grounding}: Interacting with software requires knowledge of the environment, such as the layout of elements on the screen. For automatic interaction with web software,we propose to extract and filter pertinent information from the DOM elements as the environmental observations.

\item \textbf{Sequential Decision-Making}: As the environment responds dynamically to actions, we take a sequential approach to generate actions rather than creating an entire actions sequence all at once. This allows us to generate actions informed by current observations while maintaining previous action-observation pairs. 

\end{itemize}

An additional challenge is how to train the LLM to better control software. This study limits the focus to in-context learning. One option is to manually create examples of programs that satisfy instructions, but this requires some level of expertise and experimentation to be effective. Observing that the LLM occasionally produces successful programs for specific tasks (rather than always or never), we propose using these successes as a form of context. This way, the LLM can create its own context using only a success indicator.

In this paper, our \textbf{main contribution} is to investigate the use of an LLM to control web software, particularly:
\begin{itemize}
    \item Use of filtered DOM elements as web software observations, which we find outperforms more comprehensive read-outs.
    \item Effects of an iterative approach of cycling action and observation, which we show outperforms one-shot generation of an action sequence. 
    \item Auto-generating context based on successful zero-shot trials, which outperforms zero-shot performance in many cases and requires no programming or knowledge of the control API. 
\end{itemize}

\section{Related Work}
\subsection{Automated Computer Tasks} 
Most works for automating computer tasks use some form of reinforcement learning. Common approaches, including Q-Learning ~\citep{jia2018domqnet}, imitation learning~\citep{yao2022webshop}, and policy learning behavioral cloning~\citep{zheng2021automatic} have achieved human-level performance on the web interface benchmark~MiniWob++~\citep{zheran2018reinforcement}. Other approaches, such as~\citep{pmlr-v162-humphreys22a,zhong2022improving} combine RL with other modalities. CCNet~\citep{pmlr-v162-humphreys22a} is a multi-modal architecture specifically designed for automating software, which is trained using RL, imitation learning, and also both.  Language Dynamics Distillation~\citep{zhong2022improving} is pre-trained by predicting environment information and fine-tuned with RL. 
While these methods can work well, they often require thousands of demonstrations and/or millions of trials.  We investigate the use of LLM with only very limited in-context learning, towards creating an approach that is more easily extended and adopted.

\subsection{Reasoning and Action in Large Language Models}
Recent works explore applying LLMs to decision-making~\citep{mialon2023augmented} and reasoning~\citep{huang-chang-2023-towards}. Initial work focused on how to convert natural language output to admissible actions~\citep{huang2022language}. Huang et al. use a Bert LM~\citep{devlin2019bert} model pre-trained with SentenceBert~\citep{sentencebert} to directly convert the output of GPT3~\citep{NEURIPS2020_1457c0d6} to an executable action. SayCan~\citep{ahn2022can} uses an alternative approach where for each action, the probability of generating that action in a language model is multiplied by the action's value function. Inner Monologue~\citep{huang2022inner} builds on SayCan~\citep{ahn2022can} by introducing feedback from the environment. Unlike these approaches, our web environment is much simpler, and we found that with specific prompting, we can convert natural language output to an appropriate action. Prompting, in particular, chain-of-thought prompting has been used to demonstrate LLM's reasoning ability~\citep{wei2022chain,kojima2022large,nye2021show}, and an emerging area of research is combining reasoning and decision-making skills into one method. SayCan~\citep{ahn2022can} and Inner Monologue~\citep{huang2022inner} are both early examples of such a combination. ReAct~\citep{yao2023react} builds upon these works by adding language (generated by an LLM) to the list of possible actions. 
After each action is executed, a thought (language action) is generated based on the previous action and environment.  Reflexion~\citep{shinn2023reflexion} builds upon ReAct~\citep{yao2023react} by allowing access to previous actions and states. 
\subsection{Visual Programming}
Another application of LLMs is program generation~\citep{chen2021evaluating_llms_trained_on_code_codex}. VisProg~\citep{Gupta_2023_CVPR} introduced the idea of visual programming, where programs call APIs to interpret and transform images using pre-trained models, solving tasks like image editing and Visual Question Answering (VQA). VisProg~\citep{Gupta_2023_CVPR} and concurrent works~\citep{suris2023vipergpt, wu2023visual, gao2023pal, wang2022code4struct} provide a prompt to the LLM that contains the API, a set of example programs, and an instruction.  Visual programming methods produce impressive zero-shot results but are limited in that they generate one-shot programs without observing the image. Our approach generates programs in multiple steps and uses DOM elements to summarize the visual input. The method 'Recursively Criticizes and Improves' \citep{kim2023language}, a contemporaneous study, similarly employs LLMs for software interaction. Yet, their approach takes HTML as observations and incorporates multiple in-context examples.


\section{MiniWob++ Benchmark}
\begin{figure}[ht]
\centering
\includegraphics[width=0.8\textwidth]{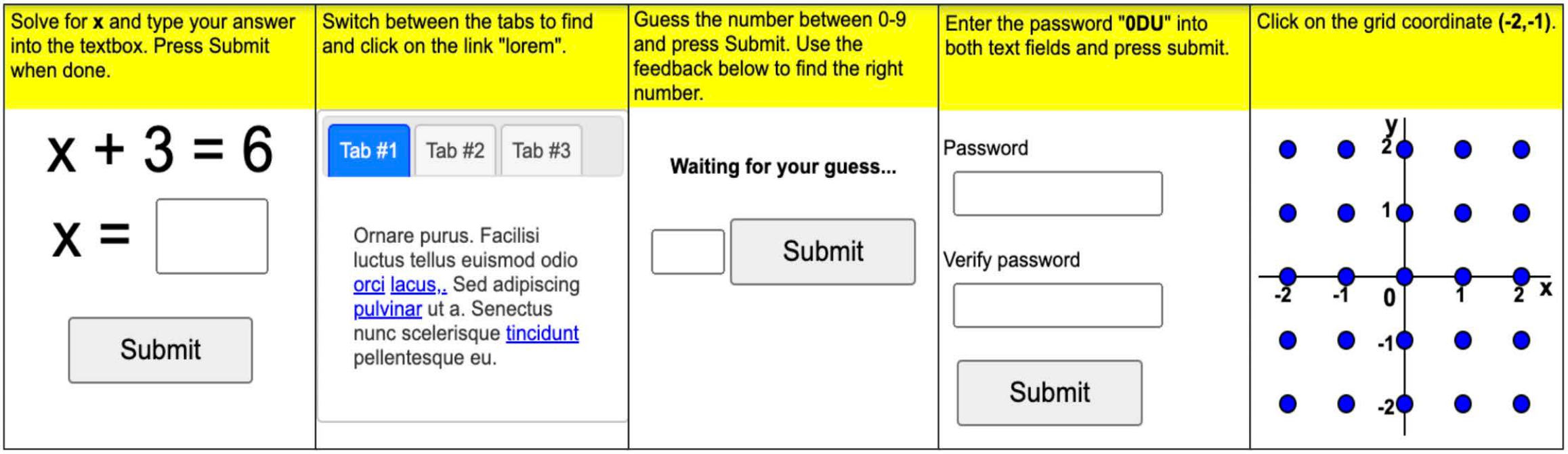}\hfill
\caption{Screenshots of example tasks from  MiniWob++ benchmark~\citep{zheran2018reinforcement}. Each task contains a natural language instruction ( with background in yellow) at the top and a task interface to perform primitive actions in the bottom (with background in white).}
\label{fig:wob_task}
\end{figure}

Our objective is to generate programs to control web interaction environments. To evaluate the effectiveness of these generated programs, we use the MiniWob++~\citep{zheran2018reinforcement} benchmark that captures the salient challenges of browser interactions in a simple setting to evaluate the generated programs. This web-based simulation environment is an extension of MiniWob~\citep{pmlr-v70-shi17a} originally introduced by OpenAI. The MiniWob++ benchmark comprises over 100 tasks of varying difficulty levels, and we chose 48 of those tasks to evaluate our methods on. Each MiniWob++ task contains a natural language instruction and an HTML/DOM representation of the web page containing the task. An agent can interact with MiniWob++ through the Selenium API. Successfully completed tasks receive a score of one. We describe the actions we use in the Methods section. A few of the tasks we selected for this study are shown in Figure \ref{fig:wob_task}.


\vspace{-4pt}
\section{Methods}
\begin{figure}[ht]
  \centering
  \includegraphics[width=\textwidth]{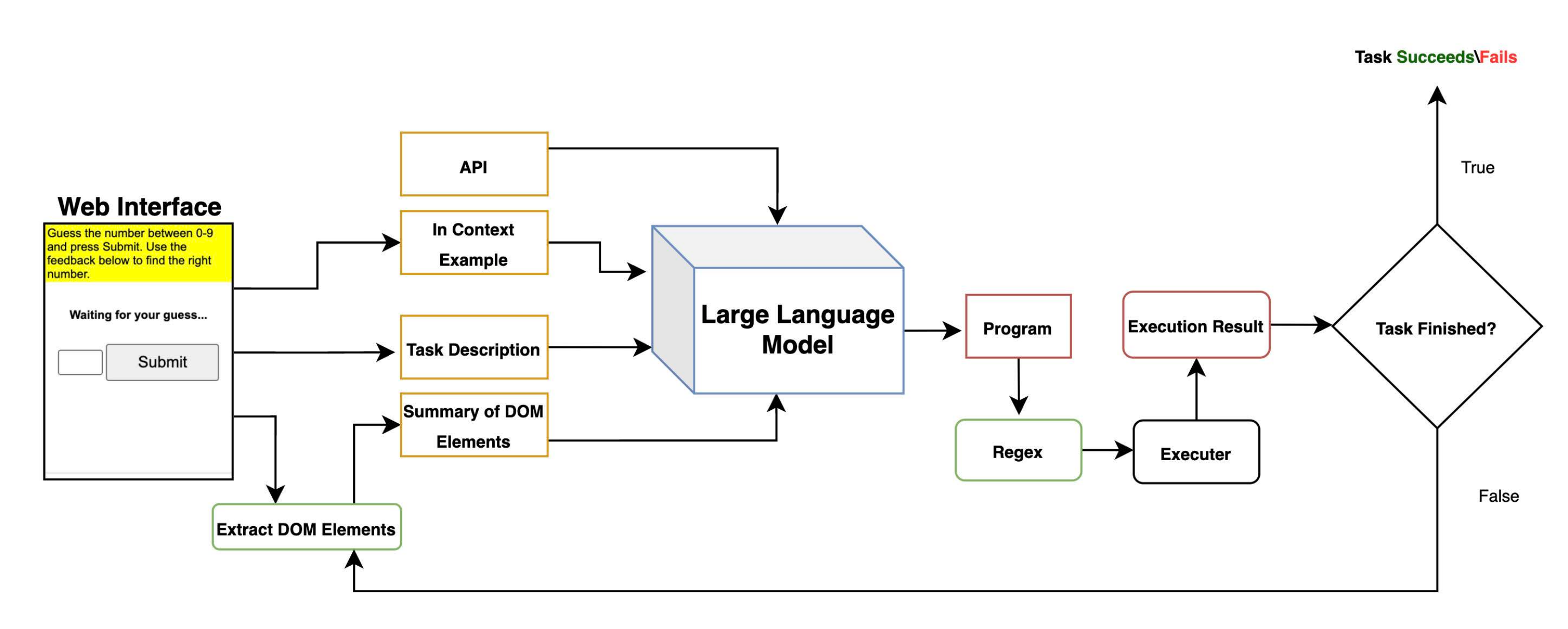}
  \caption{We show the path of an agent in the WebWISE method. The representation of the components in the diagram is as follows: yellow boxes represent inputs, the blue box represents Large Language Model (LLM), red boxes symbolize outputs, and green boxes denote Python functions. }
  \label{fig:algorithm}
\end{figure}

This section describes our problem formulation and progressive layers to our approach: based on instruction alone, incorporating filtered DOM elements as observations, acting step-by-step, and auto-generating context. See Figure~\ref{fig:algorithm} and Algorithm~\ref{alg:webwise} for an overview of our full approach, which is called WebWISE.

\begin{algorithm}
\caption{WebWISE Function}
\label{alg:webwise}
\begin{algorithmic}
\Procedure{WebWISE}{\textbf{input} Task Description $nl$, API $a$, $m$ in-context examples $\mathbf{e}=e_1,\dots,e_m$, program generator $\pi$, DOM elements $D_e$ }
    \State Initialize success state \( s \) as False, iteration count \( \textit{iter} \) as 0, maximum iterations as \( \textit{maxIter} \)
    \While{$s=False$ \textbf{and} \( \textit{iter} < \textit{maxIter} \)}
        \State $d_{t}=filterDOM(D_e)$ \Comment{Extract relevant DOM elements}
        \State $x_{t}=\pi(nl,d_{t},a,\mathbf{e})$ \Comment{Generate program}
        \State $s = \phi(x_t,d_t)$ \Comment{Execute program}
        \State \( \textit{iter} = \textit{iter} + 1 \)
    \EndWhile 

\EndProcedure 

\end{algorithmic}
\end{algorithm}





\subsection{Problem Formulation}
The scenario under consideration involves a virtual agent assigned to carry out high-level natural language instructions (task description) denoted as $nl$. The agent must interact with the environment, which is initially represented by a set of observable DOM Elements $d_{t}$. A program generator $\pi$ is employed to generate program $x_{t}$ = $\pi(nl,d_{t})$. The execution engine $\phi$ then applies the generated program through $\phi(x_{t},d_{t})$, resulting in success indicator $s$ and updated DOM Elements $d_{t+1}$, representing the environment's next state. A maximum number of iterations $maxIter$ has also been set to limit the total number of iterations that the LLM can generate programs.

\subsection{DOM Elements}
The choice between DOM elements, HTML, and RGB values when interacting with a web interface largely hinges on the task's demands, the intricacy of the webpage, and the agent's capabilities. In this work, DOM elements were used due to their simplicity and structured nature. In situations where there's no direct access to the DOM elements, leveraging pre-trained models like Pix2Struct \citep{lee2023pix2struct} to derive structured representations (HTML/DOM) from webpage screenshots might serve as a viable approach.  The DOM elements allow the agent to engage directly with the page -- clicking on elements, inputting text into form fields, reading text from the page, and so forth -- and is generally simple to implement and comprehend.  The full text of the DOM elements can contain items not crucial for a particular task, so we use a simple filtering function $filterDOM$ that returns a subset of the current DOM elements that belong to a pre-defined list of "tags" and "classes". Further details are in the supplementary.



\subsection{Single-Step Approach}
\begin{figure}[ht]
  \centering
  \includegraphics[scale=0.45]{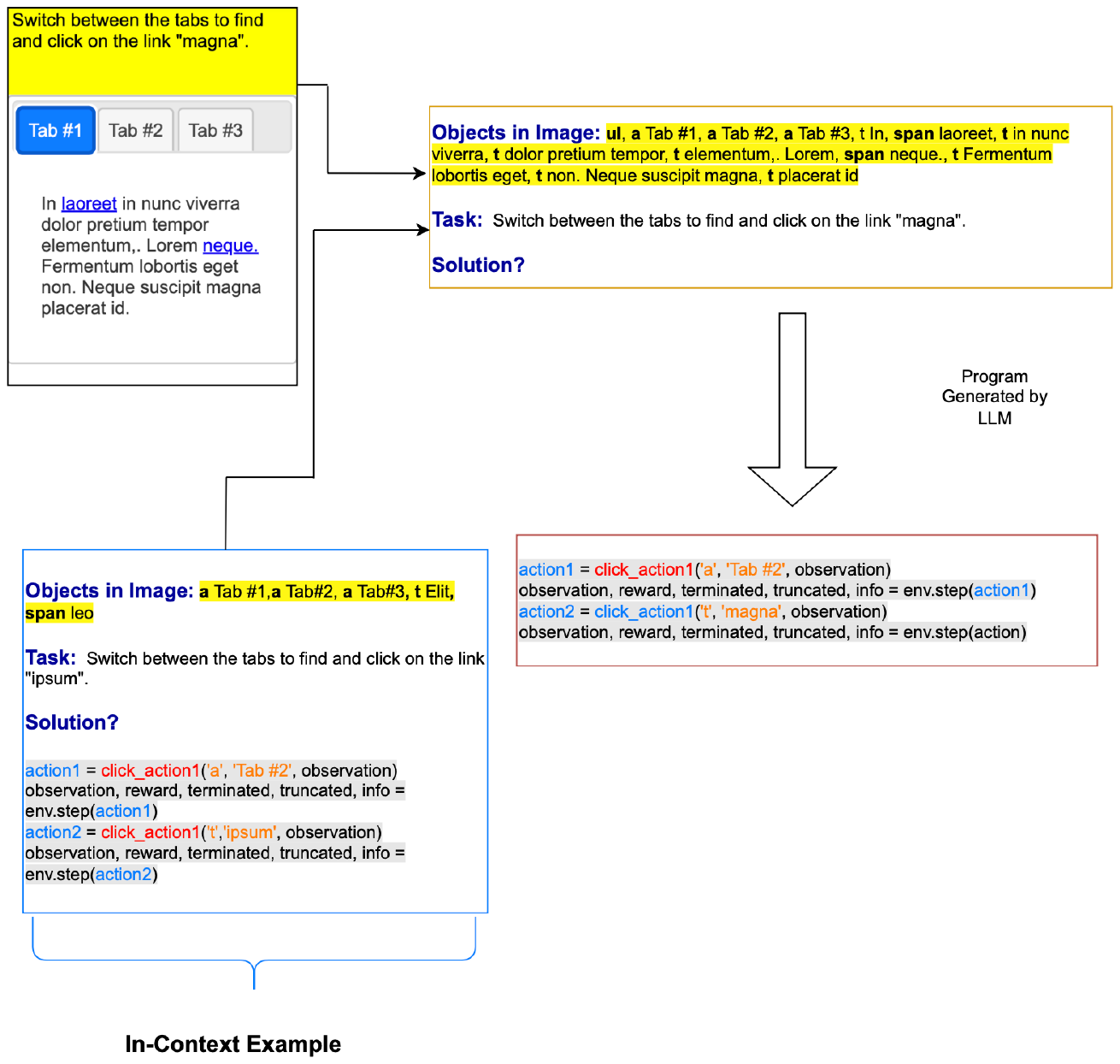}
  \caption{An example of the single-step approach. The input to the LLM includes the in-context example (in the blue box) and the task query (in the orange box). }
  \label{fig:text+dom}
\end{figure}

In the single-step approach, one program is generated and executed for a given task. A popular LLM (while working on the paper), gpt-3.5-turbo~\cite{ouyang2022training} was used as our program generator $\pi$. The input to the LLM is a prompt with 4 parts: filtered DOM elements, API $a$, task description, and in-context example $e$. Our API $a$ has three basic functions: click, enter text, and scroll. Each executes actions within MiniWob++~\citep{zheran2018reinforcement}. A summary of our API can be seen in Listing~\ref{lst:api1} and is constant across all tasks. The full version can be found in the supplementary. Hand-crafted in-context examples were used like GPT-3~\cite{NEURIPS2020_1457c0d6}, which can be seen in Figure~\ref{fig:text+dom}. A sample task input is given followed by the expected output program. Note that the input only contains the filtered DOM elements and not the entire DOM elements. 

The generated program, $x_t$ is a string of Python code. It is executed using the execution engine $\phi$. However, the code generated might not be executable or use the API $a$ correctly. A regular expression is used to extract the executable code and filter the irrelevant parts of the generated output. If the generated code proves entirely unusable, this particular program is labeled as unsolvable automatically, and LLM will move to the next iteration of generating programs to solve this task. If usable code is identified, $\phi(x_t, d_t)$ will be executed and LLM will rely on the environmental feedback to determine whether the task has been successfully completed.  The iterations continue until either the task is successfully completed (i.e., \(s = \text{True}\)) or the number of iterations reaches the predefined maximum (\(\textit{iter} = \textit{maxIter}\)). If the maximum number of iterations is reached without success, the algorithm terminates without finding a solution.

\begin{lstlisting}[language=Python,basicstyle=\tiny, caption=Summary of our API,label={lst:api1},belowskip=-0.8 \baselineskip]
def getSummary(dom_elements):
    ''' 
    Input: DOM elements
    Output: Subset of DOM elements
    '''
def click_action1(tag_class_name,id_text_name,observation):
    '''
    Input: tag or element, id or text, observation
    Output: clicks on a specific element in the environment
    '''
def enter_text_action(input_text,observation):
    '''
    Input: text, observation
    Output: enters text in element in the environment
    '''
def scroll_action1(text_to_scroll_to,observation):
    '''
    Input: text, observation
    Output: moves webpage such that certain text is visible
    '''
\end{lstlisting}

\subsection{Multi-Step Approach}
Our multi-step methodology, referred to as WebWISE, is illustrated in Algorithm \ref{alg:webwise}. In contrast to a single-step generation, WebWISE generates and executes programs incrementally until the environment signals that the task has been solved correctly or incorrectly. If the task has not been solved after a particular number of iterations, the environment signals a failure. The Document Object Model (DOM) elements from the $i$th iteration becomes the initial DOM elements for the $(i+1)$th iteration. This approach is employed in both zero-shot and one-shot scenarios. In the zero-shot scenario, an additional instruction ($task\ message$) is provided to the Large Language Model (LLM), prompting it to generate programs progressively, based on alterations in the DOM elements throughout the task. In the one-shot scenario, the in-context examples shift at each step, coupled with the additional instruction ($task\  message$) to perform in a step-by-step fashion, akin to the zero-shot scenario. Ablation studies concerning the sensitivity of the $task\ message$ are included in the supplementary section. Figure \ref{fig:webwise} illustrates an example of the WebWISE methodology. Conditioning the program generation on the current environment simplifies the execution of sequential tasks in a step-by-step manner. This is because the model can comprehend the effect of its actions in the environment, and consequently generate more effective actions for the next step. In contrast, the single-step approach would generate the entire program based solely on the initial set of DOM elements. 

\subsection{Automatic Generation of In-Context Examples From Scored Trials}
The empirical findings reveal improved success rates when an in-context example is provided. This includes an observation, task description, and a program that would satisfactorily complete the task according to the task description. However, supplying such examples necessitates a comprehensive understanding of the API and programming proficiency, which can pose an obstacle for intricate tasks or novice users. Moreover, creating these examples for web interface tasks can be costly and time-consuming due to the complexity of some tasks. Prompted by these challenges, a simple approach termed Auto-Context Generation (ACG) was designed to autonomously generate an in-context example from a successful trial. A series of zero-shot trials were conducted. In each trial, a program was generated and executed in the absence of any in-context examples. If the program executes the task correctly, it will be preserved, along with the original task description and filtered DOM elements, as the in-context example. After a set of 10 zero-shot trials, the single-step approach is then applied, as previously, for 50 iterations. The correctly generated programs during the trial stage serve as in-context examples. Only two such programs are preserved at maximum. When appending the in-context examples to the prompt, a specific statement:"Here is one example you have solved with a successful solution." is also included. While our experiments use this fully automated approach, a user, in a practical context, may be able to guide the successful completion using feedback and prompts to acquire the in-context example without the need for programming. 

\begin{figure}[ht]
  \centering
  \includegraphics[scale=0.51]{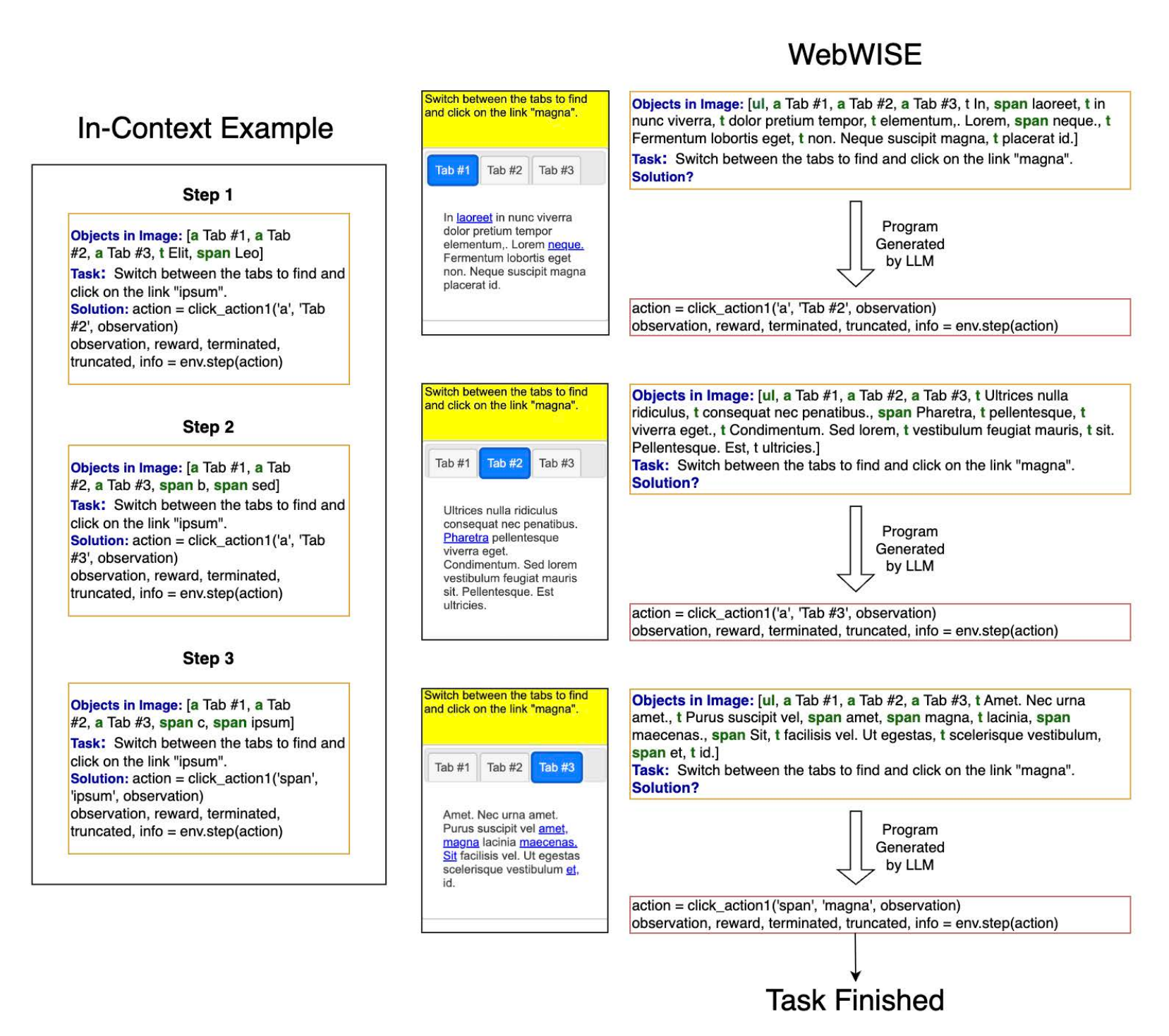}
  \caption{An example of WebWISE for the `click-tab' task. The goal is to click on a link that is not visible in the initial set of DOM elements. On each step, a context based on the updated environment is used to generate a new program. The black box on the left part of the image is the step-by-step in-context example with made-up DOM elements and tasks. This example is provided at each step of the LLM generating program based on the current task interface. Notice that in this example, one action is executed at a time. The middle and right part of the image are the actual task provided in the MiniWob++ benchmark and the solution generated by the LLM.}
  \label{fig:webwise}
\end{figure}

\vspace{-6pt}
\section{Experiments and Results}

The evaluation of the single-step, multi-step, and Auto-Context Generation techniques is conducted across 48 tasks derived from the MiniWob++ dataset~\citep{zheran2018reinforcement}. These tasks are meticulously chosen to span a range of complexities. Additionally, a variant of the single-step method, which excludes any DOM elements, is explored and labeled as the 'Instruction Only' method. The conventional single-step process is termed the 'Instruction+ Filtered DOM' approach.
 
\subsection{Implementation details}
For all experimental results, we employed the gpt-3.5-turbo model~\cite{ouyang2022training}, with a temperature parameter set to 0 and a maximum input token limit of 4096. A consistent introduction prompt is applied prior to each method, which can be found in the supplementary material. We conducted evaluations for each task using all the methods, both with and without an in-context example, denoted as $k$. Tasks were scored as follows: 1 for successful completion and -1 for failure. For the `Instruction-Only', `Instruction+DOM', and `WebWISE' (multi-step) approaches, we executed each task over 50 iterations and calculated the final accuracy accordingly. For the `Auto-Context Generation' method, we initially carried out 10 zero-shot trials, followed by 50 iterations of the `Instruction+DOM' method.

We categorized various tasks from the MiniWob++ dataset~\citep{zheran2018reinforcement} based on the number of predefined function calls necessary to accomplish the task. The groups include tasks requiring 1 function, 2 functions, between 3 and 6 functions, and a variable number of functions. A more detailed task classification is available in the supplementary material.

\subsection{Results}
We present Table~\ref{tab:results} summarizing the average success rate of our proposed methods and compare with other prior reinforcement learning (RL) and behavior cloning (BC) based approaches. Results from prior works were grouped into the same categories as our tasks. For simple tasks which requires just one function call , we outperform WebNT5-3B benchmark by a slight margin of $14.1\%$ which employs a finetuned large language model with 12K expert demonstration data.  We  outperform WebN-T5-3B, CCNet (RL), and CCNet (BC). CCNet (BC+RL) significantly outperforms our approach, but requires many expert demonstrations and millions of RL trials, while our approach requires minimal per-task learning. 

\subsubsection{Ablation}
The top half of Table \ref{tab:results} compares our proposed methods across the task groups for both zero-shot(k=0) and single-shot(k=1). For a particular value of k, Instruction Only has the lowest performance though the gap between Instruction Only and the other methods grows as the task become more complex (left to right). For the easiest tasks, using DOM elements and feedback (WebWISE) has little effect on the performance. For k=1, WebWISE produces the largest gain in the Variable Function group. Auto-context generation has a similar or higher performance than the other zero-shot methods but it is lower than the single-shot ones.

\begin{table}[ht]
\centering
\adjustbox{max width=\textwidth}{%
\begin{tabularx}{\linewidth}{c*{7}{X}X}
\toprule
\multirow{2}{*}{Methods} & \multicolumn{2}{c}{1 Function} & \multicolumn{2}{c}{2 Function} & \multicolumn{2}{c}{3-6 Function} & \multicolumn{2}{c}{Variable Function} \\
\cmidrule(lr){2-3} \cmidrule(lr){4-5} \cmidrule(lr){6-7} \cmidrule(lr){8-9}
& k=0 & k=1 & k=0 & k=1& k=0 & k=1 & k=0 & k=1\\
\midrule
Instruction Only & .45 & .94 & .03 & .27 & 0 & .94 & .01 & .20 \\
Instruction + Filtered DOM & .86 & .97 & .64 & .91 & .80& .97 & .21 & .44 \\
Instruction + Whole DOM & .82 & .94 & .71 & .89 & .77 & .80 & .16 & .23 \\
WebWISE & .87 & .97 & .73  & .93 & .80 & .99 & .27 & .75 \\
Auto-Context Generation & .86 & - &  .81& -& .80 & - & .30 & - \\
\midrule
WebNT5-3B~\citep{gur2022understanding} & \multicolumn{2}{c}{0.83} & \multicolumn{2}{c}{0.29} & \multicolumn{2}{c}{0.73}&  \multicolumn{2}{c}{0.37}\\
WebN-T5-3B(k=0) & \multicolumn{2}{c}{0.85} & \multicolumn{2}{c}{0.27} & \multicolumn{2}{c}{0.63}&  \multicolumn{2}{c}{0.30}\\
CCNet(BC+RL)~\citep{pmlr-v162-humphreys22a} & \multicolumn{2}{c}{0.99} & \multicolumn{2}{c}{0.94} & \multicolumn{2}{c}{0.99}&  \multicolumn{2}{c}{0.89}\\
CCNet(RL)~\citep{pmlr-v162-humphreys22a} & \multicolumn{2}{c}{0.88} & \multicolumn{2}{c}{0.65} & \multicolumn{2}{c}{0.50}&  \multicolumn{2}{c}{0.44}\\
CCNet(BC)~\citep{pmlr-v162-humphreys22a} & \multicolumn{2}{c}{0.77} & \multicolumn{2}{c}{0.37} & \multicolumn{2}{c}{0.27}&  \multicolumn{2}{c}{0.16}\\
WebGUM(HTML)~\citep{furuta2023multimodal} & \multicolumn{2}{c}{0.92} & \multicolumn{2}{c}{0.40} & \multicolumn{2}{c}{1.00}&  \multicolumn{2}{c}{0.83}\\
WebGUM(HTML+Image)~\citep{furuta2023multimodal} & \multicolumn{2}{c}{0.94} & \multicolumn{2}{c}{0.40} & \multicolumn{2}{c}{1.00}&  \multicolumn{2}{c}{0.90}\\
\bottomrule
\end{tabularx}}
\caption{Average success rate across tasks. Versions of our approach are in the first four rows, with WebWISE and Auto-Context Generation (ACG) being the main approach. k=0 means no in-context example is provided, or in case of ACG, only auto-context examples are provided; k=1 indicates one manual in-context example is provided, though many examples are shared across tasks. The remaining are reported from other works where the value of k is not applicable. CCNet has been trained using behavior cloning (BC) on human-labeled data and also reinforcement learning by interacting with MiniWob++ different tasks environment. CCNet (BC) represents the model that has been only trained on human-labeled data, CCNet (RL) is trained only by letting it interacting with MiniWob++ environment for many trials, and CCNet (BC+RL) is trained using both methods. }
\label{tab:results}

\end{table}

\vspace{-12pt}
\section{Discussion}


\vspace{-6pt}
\begin{figure}[ht]
     \centering
     \begin{subfigure}[b]{0.45\textwidth}
         \centering
         \includegraphics[width=\textwidth]{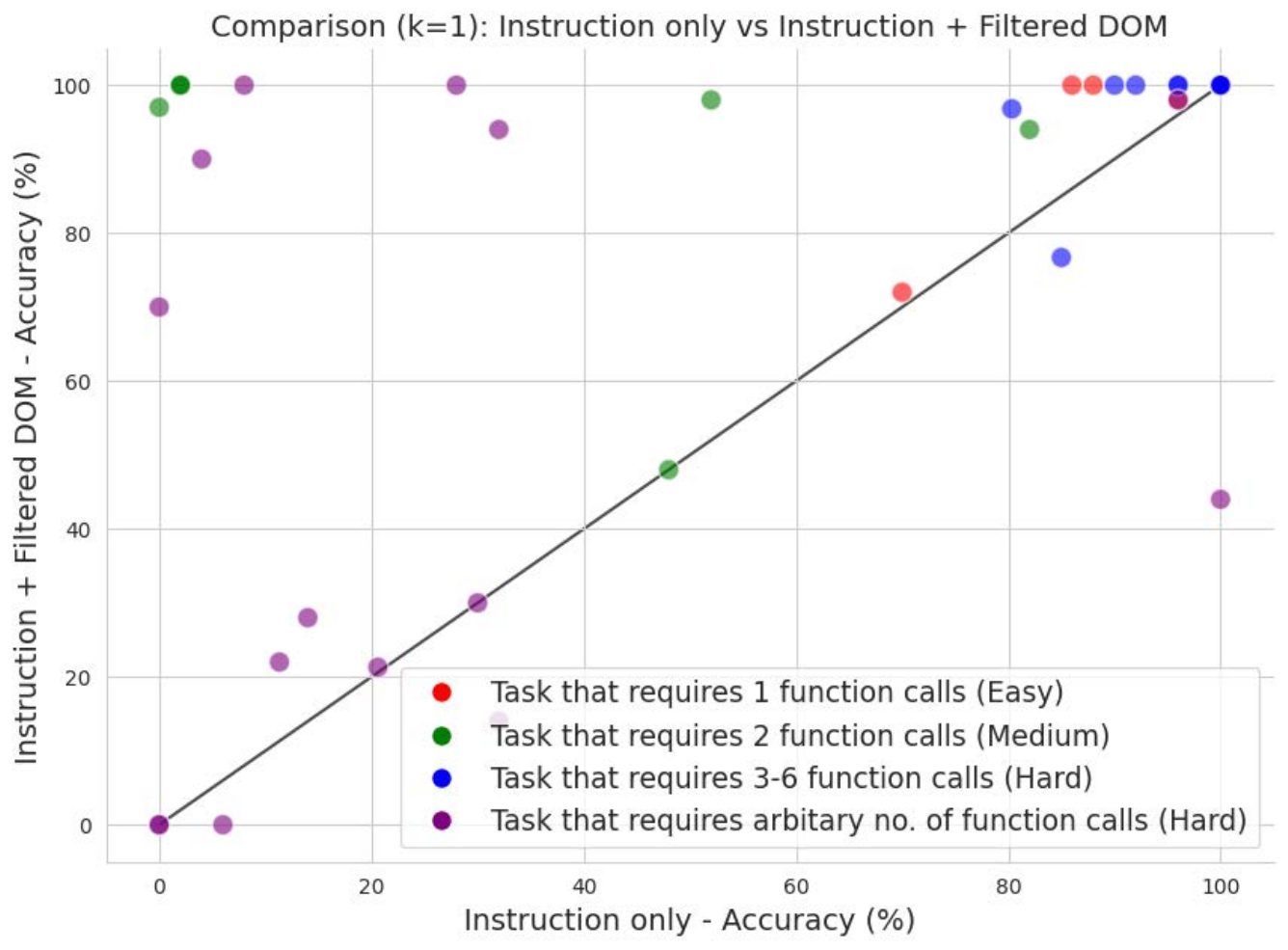}
         \caption{Comparison of k=1 performance across all tasks for Instruction Only (x-axis) and Instruction+Filtered DOM (y-axis)}
         \label{instruct_instruct_dom}
     \end{subfigure}
     \hfill
     \begin{subfigure}[b]{0.45\textwidth}
         \centering
         \includegraphics[width=\textwidth]{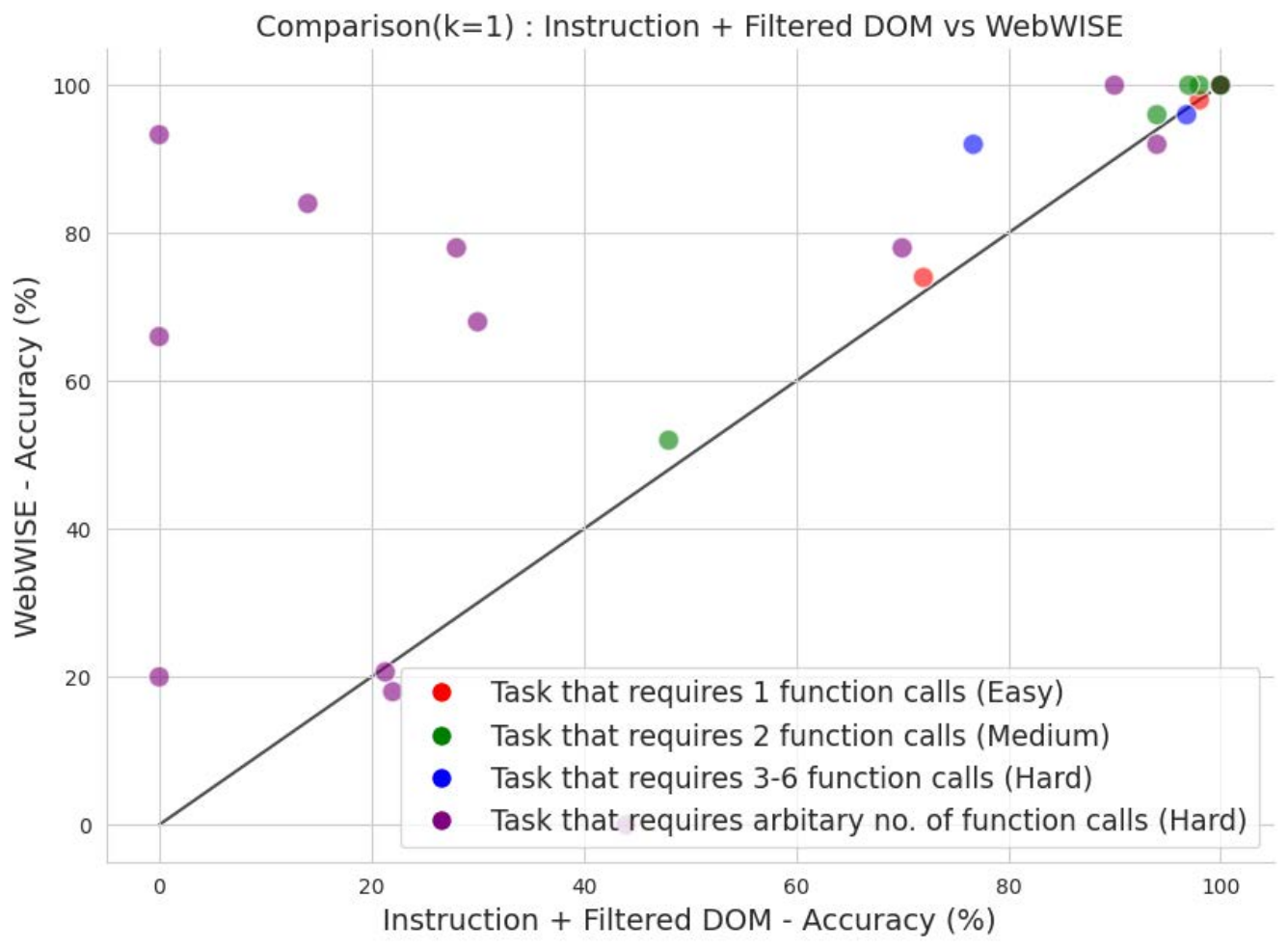}
         \caption{Comparison of k=1 performance across all tasks for Instruction+Filtered DOM (x-axis) and  WebWISE (y-axis)}
         \label{instruct_dom_webwise}
     \end{subfigure}
     \hfill
     \begin{subfigure}[b]{0.45\textwidth}
         \centering
         \includegraphics[width=\textwidth]{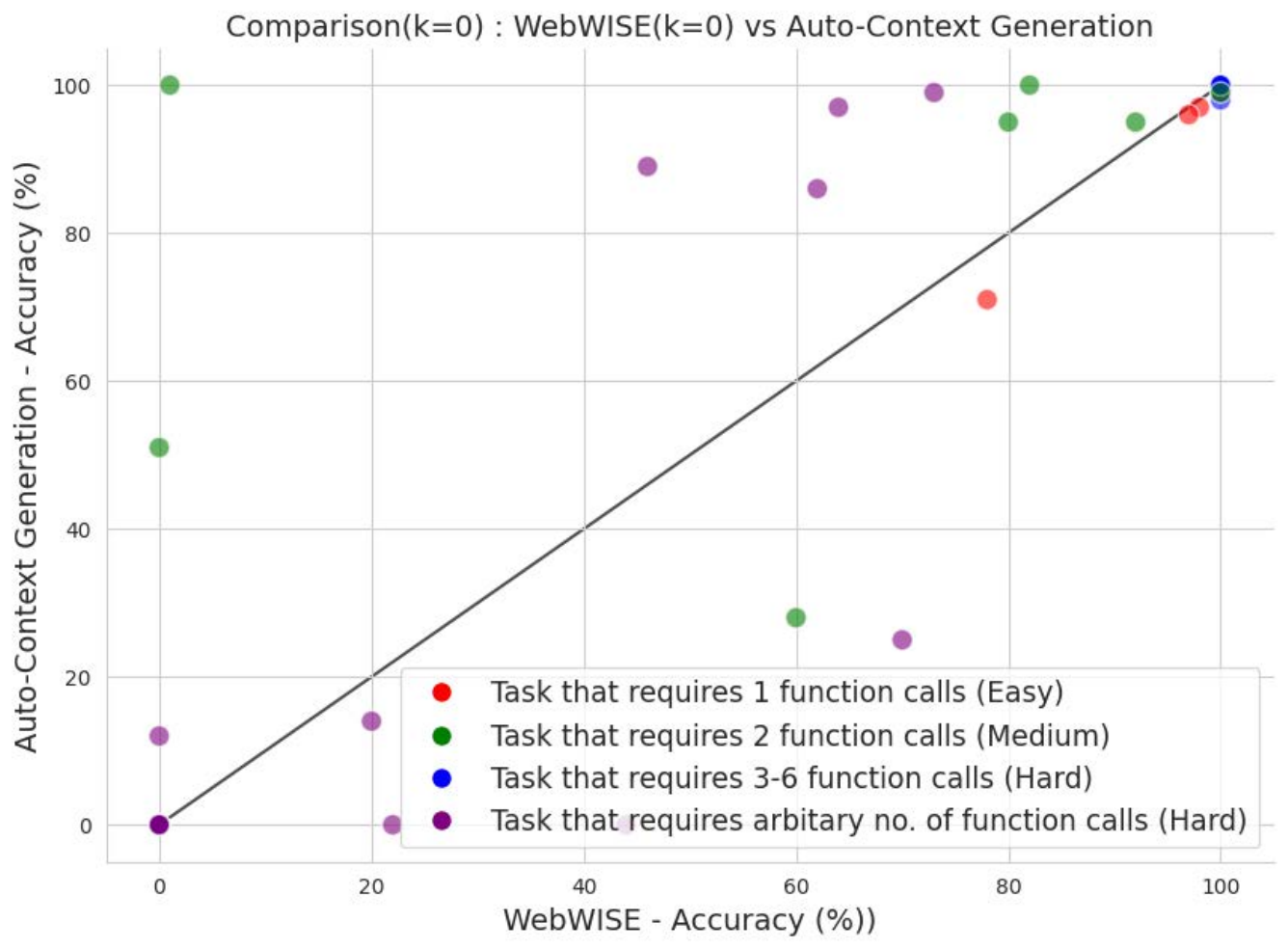}
         \caption{Comparison of k=0 performance across all tasks for WebWISE(x-axis) and  Auto-Context Generation (y-axis)}
         \label{webwise_auto_contexts}
     \end{subfigure}
    \hfill
    \begin{subfigure}[b]{0.45\textwidth}
         \centering
         \includegraphics[width=\textwidth]{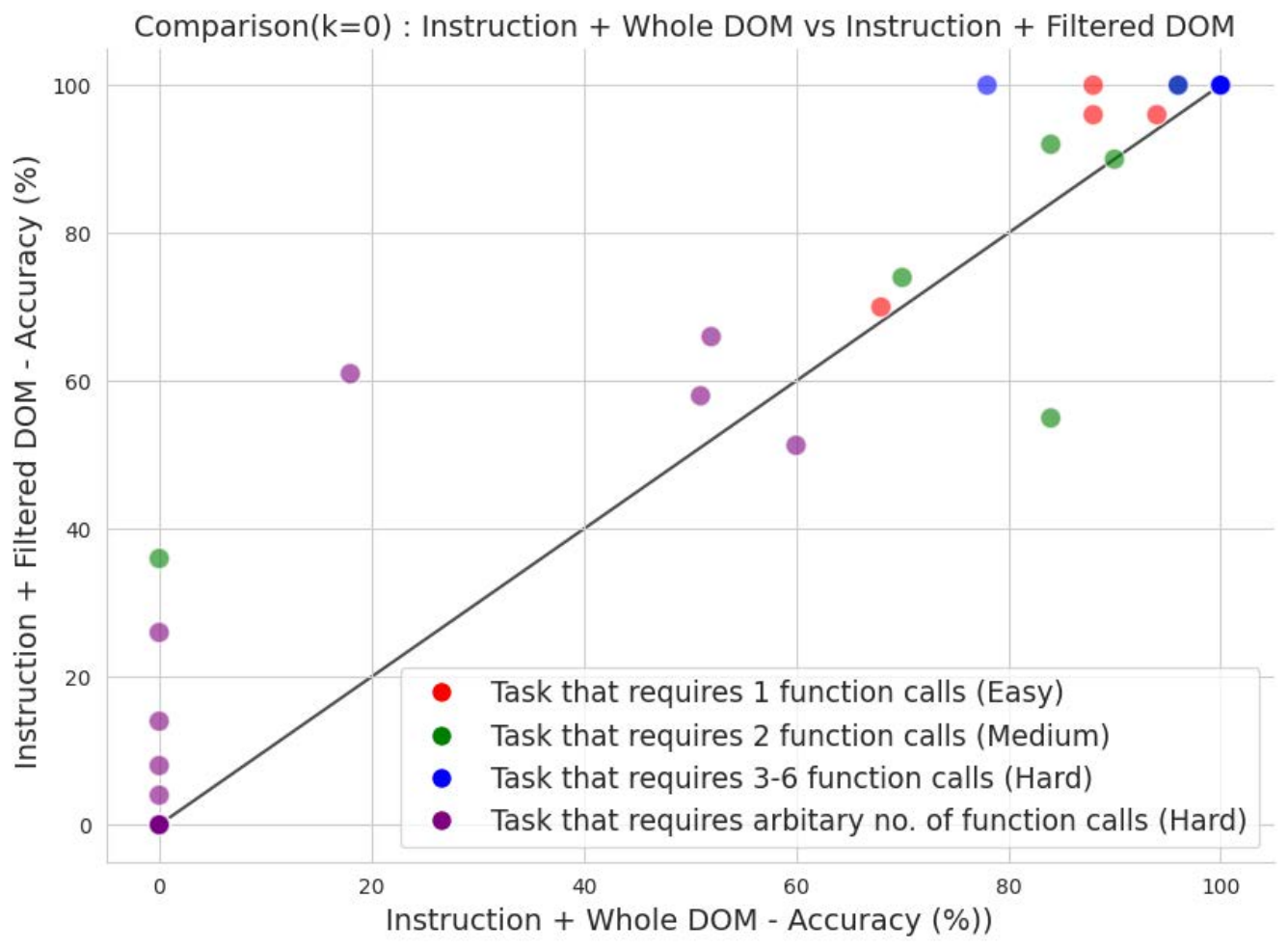}
         \caption{Comparison of k=0 performance across all tasks for Instruction+Whole DOM(x-axis) and  Instruction+Filtered DOM(y-axis)}
         \label{entire_dom_filtered_dom}
     \end{subfigure}
\caption{Scatter plots comparing performance across all tasks for different methods. }
\label{fig:scatter}
\end{figure}

 Drawing from prior work that has demonstrated the ability of LLMs in addressing various tasks in computer vision and embodied tasks, our research aims to extend the application of LLMs to more intricate challenges, specifically web interface tasks. In doing so, we shifted the focus from a reliance multiple learning examples to an approach based on zero-shot to one-shot learning, utilizing minimal examples to steer the model's learning process. We present key insights from Table~\ref{tab:results} and Figure~\ref{fig:scatter}. 

\textbf{Influence of Single In-Context Example and DOM Elements} We explored the impact of one-shot prompts, represented as k=1, across several task groups. The results indicate the ability of a LLM to perform well based on a single correct example.  As shown in Figure~\ref{instruct_instruct_dom}, most easy tasks can be successfully completed even without observations (DOM elements). However, as the task complexity increases, the inclusion of observations becomes critical which is evident from the results. 




\textbf{Improvement from Step-by-Step Actions}: \textbf{Enhancement from Sequential Actions}: Both Table \ref{tab:results} and Figure~\ref{instruct_dom_webwise} clearly illustrate that implementing actions and observations in a sequential manner (WebWISE) significantly enhances performance compared to a single-shot action generation. This improvement is especially noticeable in the most challenging task set, where many tasks necessitate the execution of actions in a specific sequence and observing the changes to refine the previous action. With the inclusion of a single in-context example (k=1), we observe a performance leap from 23\% to 75\%.


\textbf{Performance of Auto-Context Generation (ACG)}: Figure~\ref{webwise_auto_contexts} shows that auto-context examples increase success rates, compared to no in-context examples, for the large majority of cases, including all levels of difficulty.  As Table~\ref{tab:results} shows, though, a single manually provided in-context example leads to even higher success rates, especially for the hardest tasks. While ACG is comparable to the other zero-shot methods, it still lags behind the single-shot methods across different tasks. This indicates that either the in-context examples generated are not as effective as manually created one or that on some tasks, especially the more difficult ones, no in-context examples are generated. 


\textbf{Impact of filtered DOM elements}:
Figure~\ref{entire_dom_filtered_dom} shows that filtering the DOM elements improves success rates for the large majority of tasks, as the LLM is able to focus on more relevant information. However, Figure \ref{entire_dom_filtered_dom} highlights few tasks where the performance of our approach when provided with whole DOM elements is better than the filtered DOM elements. This is because there are few additional information present in the DOM whole elements which enables the model to generate better programs. Thus developing an adaptive DOM element filter that can generalize across several tasks could be a next step. 

\section{Conclusion}

Our work presents an initial exploration into using Large Language Models (LLMs) to generate programs that interact with web interfaces.  Our experiments indicate: filtered DOM elements are effective forms of observation; the step-by-step action and observation is more effective than one-shot generation; and automatically generated in-context examples from successful trials can boost success rates for many tasks.

\textbf{Limitations}: Our main limitations include that experiments are limited to simple tasks, sensitivity to the input prompt, lack of an explicit memory, and conduct experiments on only one LLM. Eventually, we aim to develop systems that can learn to perform more complicated tasks, like booking airline tickets, with few trials or demonstrations. This requires being able to more fully utilize web interfaces and retain memory of past interactions to complete long action sequences. Increased robustness to input prompts is also needed. Further improvement may be possible by learning from failures and automatically trying to correct mistakes.

\section{Acknowldgements}
This work is supported in part by ONR awards N00014-21-1-2705 and N00014-23-1-2383. We thank Tanmay Gupta (AI2) and our lab members for their valuable feedbacks.

\bibliography{main}

\newpage
\section*{Supplementary Material}

\subsection{Results for all methods across all tasks}
\begin{table}[!htbp]
\small 
\centering
\resizebox{\textwidth}{!}{
\begin{tabularx}{\textwidth}{*{10}{c}}
\toprule
&\multicolumn{9}{c}{Accuracy} \\
\midrule 
\multirow{2}{*}{Task}&\multicolumn{2}{l}{\makecell{Instruction \\ Only}}&\multicolumn{2}{l}{\makecell{Instruction+\\Filtered DOM}} & \multicolumn{2}{l}{\makecell{Instruction+\\Whole DOM}} & \multicolumn{2}{l}{WebWISE} & \multicolumn{1}{l}{\makecell{Auto-Context \\ Generation}}\\ 
\cmidrule(lr){2-3} \cmidrule(lr){4-5} \cmidrule(lr){6-7} \cmidrule(lr){8-9} \addlinespace 
&k=0&k=1&k=0&k=1&k=0&k=1&k=0&k=1&\makecell{Zero Shot\\ Trials=10}\\
\midrule 
click-button-sequence &1.00 &1.00&1.00&1.00&1.00&1.00&1.00&1.00&1.00\\
click-button &1.00 &1.00&1.00&1.00&1.00&1.00&1.00&1.00&1.00\\
click-checkboxes-large &0.00 & 0.04&0.26&0.90&TLE&TLE&0.46&1.00&0.89\\
click-checkboxes-soft &0.00&0.00&0.61&0.70&0.18&0.41&0.62&0.78&0.86\\
click-checkboxes-transfer& 0.04&0.28&0.66&1.00&0.52&TLE&0.73&1.00&0.99\\
click-checkbox&0.08&0.08&0.58&1.00&0.51&1.00&0.64&1.00&0.97\\
click-collapsible-2&0.00&0.00&0.00&0.00&0.00&0.00&0.00&0.00&0.00\\
click-collapsible&0.00&.002&1.00&1.00&0.96&1.00&1.00&1.00&1.00\\
click-dialog-2&0.64&0.70&0.70&0.72&0.68&0.54&0.78&0.74&0.71\\
click-dialog&0.00&1.00&1.00&1.00&.88&1.00&1.00&1.00&1.00\\
click-link & 0.00 & 0.88&0.96&1.00&0.88&0.92&.98&1.00&0.97\\
click-option&0.18&0.02&0.74&1.00&0.70&TLE&0.82&1.00&1.00\\
click-pie&0.00&0.00&0.00&0.00&0.00&0.00&0.74&0.80&0.52\\
click-tab-2-hard & 0.00&0.30&0.14&0.30&TLE&TLE&.22&.68&0.00\\
click-tab-2-easy & 0.00 & 0.82& 0.92& 0.94&0.84&0.86&0.92&0.96&0.95\\
click-tab-2-medium& 0.00 & 0.48 & 0.36&0.48&TLE&TLE&0.60&0.52&0.28\\
click-tab-2&0.00&0.14&0.04&0.28&TLE&TLE&0.44&0.78&0.00\\
click-tab&0.00&1.00&1.00&1.00&1.00&1.00&1.00&1.00&.99\\ 
click-test&1.00&1.00&1.00&1.00&1.00&1.00&1.00&1.00&1.00\\
click-test-transfer&1.00&1.00&1.00&1.00&1.00&1.00&1.00&1.00&1.00\\
click-test-2&1.00&1.00&1.00&1.00&1.00&1.00&1.00&1.00&1.00\\
click-widget&0.02&0.96&0.96&0.98&0.94&1.00&0.97&0.98&0.96\\
enter-date&0.00&1.00&1.00&1.00&1.00&1.00&1.00&1.00&1.00\\
enter-password&0.00&0.96&1.00&1.00&1.00&1.00&1.00&1.00&1.00\\
enter-text-dynamic&0.00&0.96&1.00&1.00&1.00&1.00&1.00&1.00&0.98\\
enter-text-2&0.00&.90&1.00&1.00&0.96&0.96&1.00&1.00&1.00\\ 
enter-text&0.00&1.00&1.00&1.00&1.00&1.00&1.00&1.00&1.00\\
enter-time&0.00&0.52&0.00&0.98&0.00&0.80&0.00&1.00&0.51\\ 
focus-text-2&0.00&1.00&1.00&1.00&0.78&1.00&1.00&1.00&1.00\\ 
focus-text&0.00&0.92&1.00&1.00&1.00&1.00&1.00&1.00&1.00\\ 
guess-number&0.00&0.32&0.08&0.14&0.00&0.00&0.20&0.84&0.14\\
login-user&0.00&0.96&1.00&1.00&1.00&1.00&1.00&1.00&1.00\\ 
grid-coordinate&0.00&0.86&0.00&1.00&0.00&TLE&0.00&1.00&0.00\\
multilayout&0.00&0.00&0.00&0.64&0.00&0.00&0.00&.78&0.00\\
read-table&0.00&0.00&0.28&0.48&0.24&0.30&0.80&0.86&0.90\\
read-table-2&0.00&0.00&0.20&0.32&0.16&0.22&0.80&0.82&0.88\\
simple-arithmetic&0.00&0.00&0.90&0.97&0.90&0.90&1.00&1.00&0.99\\ 
simple-algebra&0.00&0.02&0.55&1.00&0.84&0.90&0.80&1.00&0.95\\
navigate-tree&0.00&0.32&0.51&0.94&0.60&0.74&0.70&0.92&0.25\\
search-engine&0.00&0.06&0.00&0.00&0.00&0.00&0.00&0.20&0.00\\
find-word&0.00&1.00&0.00&0.44&0.00&0.18&0.00&0.00&0.00\\
email-inbox-forward-nl-turk&0.00&0.85&0.00&0.77&0.00&0.00&0.00&0.92&0.00\\
email-inbox-forward-nl&0.00&0.80&0.00&0.97&0.00&0.00&0.00&0.96&0.00\\
email-inbox-nl-turk&0.00&0.11&0.00&0.22&0.00&0.00&0.00&0.18&0.00\\
email-inbox&0.00&0.20&0.00&0.21&0.00&0.00&0.00&0.21&0.00\\
terminal&0.00&0.00&0.00&0.00&0.00&0.00&0.00&0.00&0.00\\
text-transform&0.00&0.00&0.00&1.00&1.00&1.00&0.00&1.00&0.00\\
use-autocomplete&0.00&0.80&0.84&0.92&0.80&0.92&0.92&0.92&0.90\\
\bottomrule
\end{tabularx}
}
\caption{Results across different methods and different tasks. TLE stands for `Token Limit Exceeded' which means we were unable to fit the entire prompt into the LLMs context window. }
\end{table}

\subsection{LLM Input}
The prompt to the LLM for each example contains (in the following order) first System message, the API, second System message, Task description, and In-context examples. There are a total of 48 different tasks and each tasks has its own in-context example. We provide the details for the system messages and API below. 
\paragraph{System Message}
The (first two) system messages were the same between tasks and across methods. A third system message is added for WebWise and Auto-Context Generation to ensure the task is completed step-by-step. 

\begin{lstlisting}[language=Python,basicstyle=\tiny, caption=First system message, label={lst:sm1},belowskip=-0.8 \baselineskip]
System_Message = """You are designed to generate programs to solve a wide range of complex web interface tasks.
You should be able to generate the program using either one or a composition of predefined action functions 
along with general python codes to solve different tasks. You should not conversate with human in any context."""
\end{lstlisting}

\begin{lstlisting}[language=Python,basicstyle=\tiny, caption=Second system message that appears after the API, label={lst:sd},belowskip=-0.8 \baselineskip]

Solution_Description="""Your task is to generate a solution for given problems based on objects in an image. Use the functions provided and follow these guidelines:

1)Construct solutions by calling functions and using Python data structures.
2)Solutions should be after the text 'Solution?'.
3)Only provide the function names without extra text in the solution.
4)Assume you can use observations without checking.
5)Don't assume additional functions or unknown information.
6)Add observation, reward, terminated, truncated, info = env.step(action) after each action.
7)Actions are independent of each other.
8)Do not add any comments , just return the code

If the task cannot be directly solved, perform a reasonable action and observe changes in the objects. 
Use your DOM Elements knowledge to understand objects in the image. 
Feel free to use Python constructs like if-else, for loop, while loop, etc., to generate the program."""

\end{lstlisting}

\begin{lstlisting}[language=Python,basicstyle=\tiny, caption=Third system message for multi-step methods, label={lst:sm2},belowskip=-0.8 \baselineskip]
task_message = """Your next task is a multi-step challenge. To successfully complete it, carefully observe and analyze the changes in user input before and after performing any action. This will help determine whether further actions are necessary. While it's important to explore and try various actions, always assess whether the task can be solved with the current user input before taking additional steps. Focus on efficiently reaching a solution without excessive exploration when a satisfactory outcome is already achievable. """
\end{lstlisting}

\paragraph{Full API}
Below we list the full details our API. The API is also mostly constant between tasks and methods. For each function, we list the expected input, output, description and the example use case of the function. The example differs across the methods and depends on what visual information is provided. For the `Instruction Only' method, there is no line that starts with `Objects in Image' since that information is not part of the method. 
\begin{lstlisting}[language=Python,basicstyle=\tiny, caption=Full API,label={lst:api2},belowskip=-0.8 \baselineskip]
You should only use the functions provided herewith in the function description. 
Here is the list for the pre-defined functions [getSummary, click_action1, enter_text_action, scroll_action1]

To use a function, please refer to the Name, Input, Output, Description of the functions, and usage examples below. Action functions should be called correctly in the solution. 
def getSummary(dom_elements):
    ''' 
    Input: DOM elements
    Output: Subset of DOM elements
    Description: get the filtered DOM elements from full DOM elements
    Example: objects_in_the_image = getSummary(dom_elements)
    '''
def click_action1(tag_class_name,id_text_name,observation):
    '''
    Input: tag or element, id or text, observation
    Output: clicks on specific element in environment
    Description: useful when you want to click on an element in the web interface. This function cannot be generalized on names. Normally first input is one of tag or element and second is test or id. The output is given as the action by calling click_action1 function or 'Cannot find in the DOM_element' if no such thing to be clicked on
    Example: Objects in Image: Button One;
             Task: Click button ONE;
             Solution: action = click_action1('button', 'ONE', observation)
                       observation, reward, terminated, truncated, info = env.step(action)

    '''
def enter_text_action(input_text,observation):
    '''
    Input: text, observation
    Output: enters text in element in environment
    Description: useful when you want to type the input_text into input text box or a similar object like input_number that can accept text given the observation of the task interface. Need to call click_action1 to click on it before calling this function. 
    Example: Objects in Image: input_text textbox; 
             Task: Type 'Hello' into textbox;
             Solution: action = click_action1('input_text', 'textbox', observation)
                       observation, reward, terminated, truncated, info = env.step(action)
                       action = enter_text_action('Hello', observation)
                       observation, reward, terminated, truncated, info = env.step(action)
    '''
def scroll_action1(text_to_scroll_to,observation):
    '''
    Input: text, observation
    Output: moves webpage such that certain text is visible
    Description: needed when elements do not appear on initial screen. Always used with other actions
    Example: Objects in Image: Button Apple 
             Task: scroll and click button Apple
             Solution: action = scroll_action1('Apple',observation)
                       observation, reward, terminated, truncated, info = env.step(action)
                       action = click_action1('button','Apple',observation)
                       observation, reward, terminated, truncated, info = env.step(action) 
    '''
\end{lstlisting}

\subsection{Filtered VS Full DOM Elements}
We discussed in our paper that we use only a subset of DOM Elements (referred to as `Filtered DOM Elements') instead of the entire set of DOM Elements. We use the $getSummary()$ function to extract the filtered DOM elements. The $getSummary()$ function iterates through each DOM element and includes a DOM element in the filtered list if the element's tag or class belongs to a pre-defined list. In addition, we include the flags, an array of binary values. The values indicate whether a certain element has been clicked on/modified on. We display the pre-defined lists, called `useful\_tag' and `useful\_classes' below. The lists were determined experimentally.

\begin{lstlisting}[language=Python,basicstyle=\tiny, caption=Predefined list of useful tags and classes, label={lst:sm3},belowskip=-0.8 \baselineskip]
useful_tag = {'button', 'text', 'input_time', 'textarea', 'polygon', 'label', 'input_password', 'rect', 'tt', 'circle', 'input_password', 'span', 'input_text', 'input_number', 'input_date', 'input_radio', 'tspan', 'input_checkbox', 't', 'button', 'h3', 'ul', 'a', 'p', 'div', 'th','tr','td'}

useful_classes = {'folder'}
\end{lstlisting}

\begin{figure}[ht]
  \centering
  \includegraphics[scale=0.57]{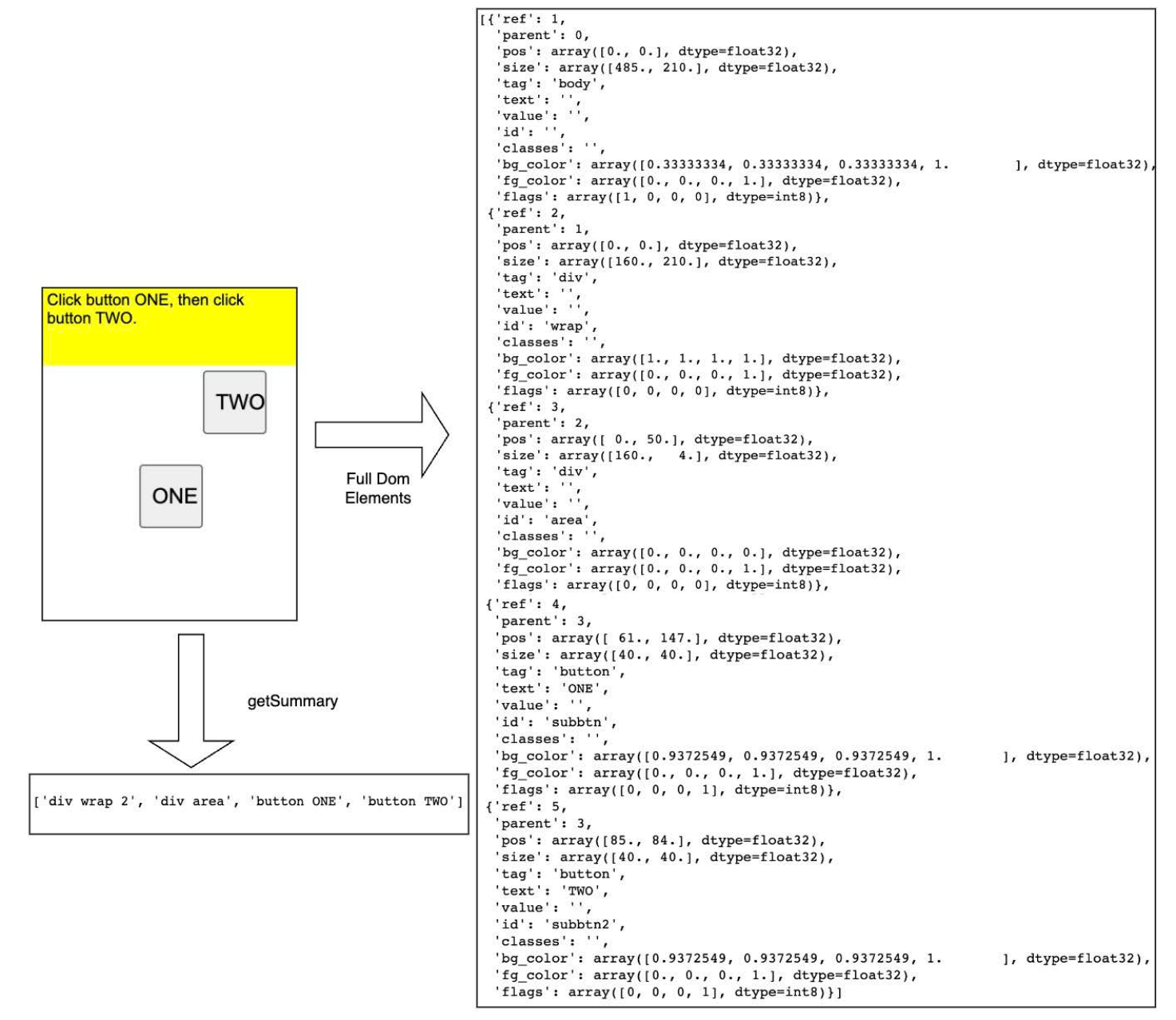}
  \caption{An example of comparing the full DOM elements VS filtered DOM elements for a simple task. We also include flags as part of the $getSummary()$ output.}
  \label{fig:filter vs full}
\end{figure}

In Figure \ref{fig:filter vs full}, we use a simple task to illustrate the differences between filtered DOM elements and full DOM elements.

\subsection{Scatter Plots}
\begin{figure}[ht]
     \centering
     \begin{subfigure}[b]{0.45\textwidth}
         \centering
         \includegraphics[width=\textwidth]{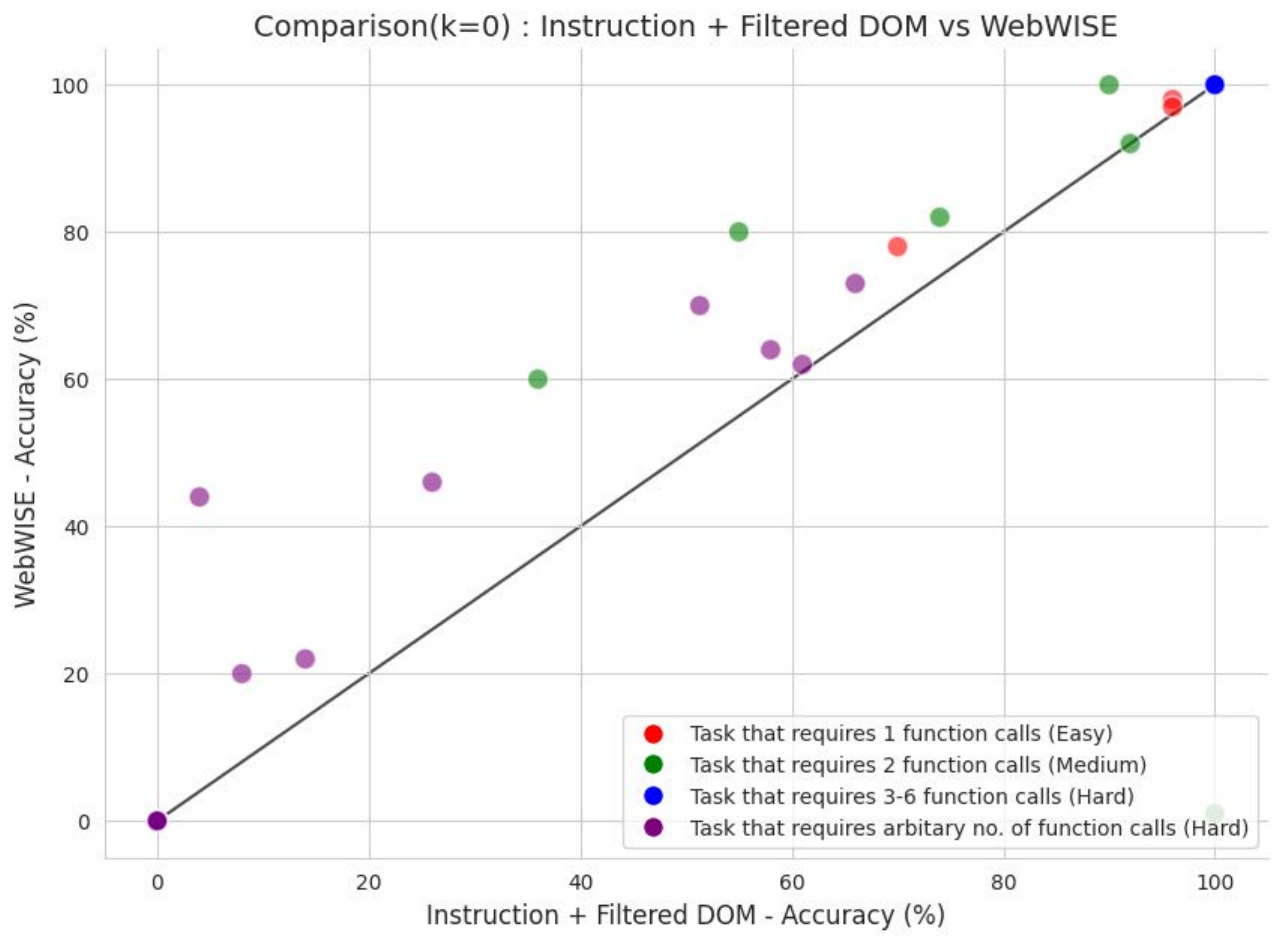}
         \caption{Comparison of k=0 performance across all tasks for Instruction Only (x-axis) and Instruction+Filtered DOM (y-axis)}
         \label{fig:sc1}
     \end{subfigure}
     \hfill
     \begin{subfigure}[b]{0.45\textwidth}
         \centering
         \includegraphics[width=\textwidth]{Images/k023.pdf}
         \caption{Comparison of k=0 performance across all tasks for Instruction+Filtered DOM (x-axis) and  WebWISE (y-axis).}
         \label{fig:sc2}
     \end{subfigure}
     \hfill
     \begin{subfigure}[t]{0.45\textwidth}
         \centering
         \includegraphics[width=\textwidth]{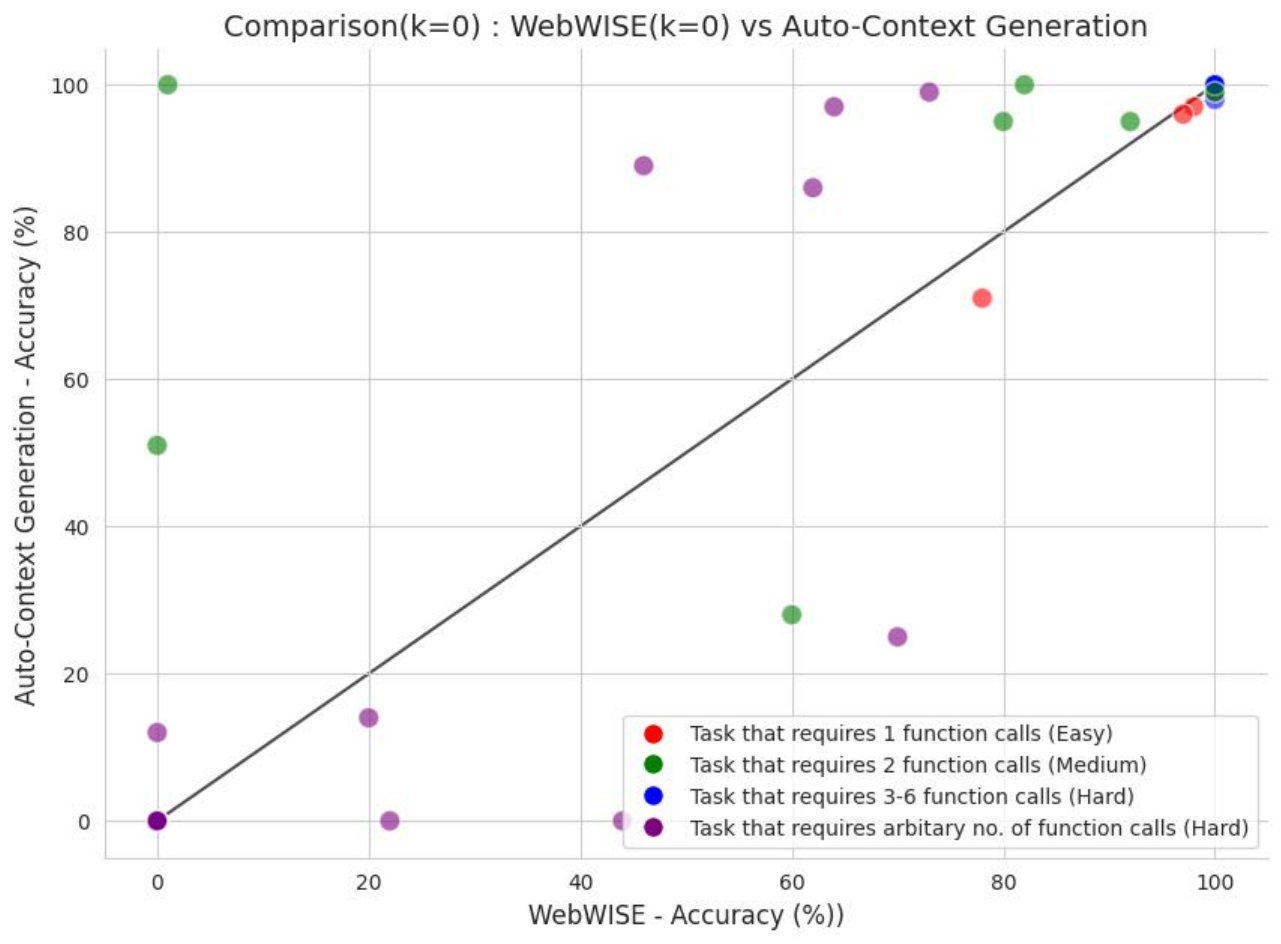}
         \caption{Comparison of k=0 performance across all tasks for WebWISE(x-axis) and  Auto-Context Generation (y-axis)}
         \label{fig:sc3}
     \end{subfigure}
     \hfill
     \begin{subfigure}[t]{0.45\textwidth}
         \centering
         \includegraphics[width=\textwidth]{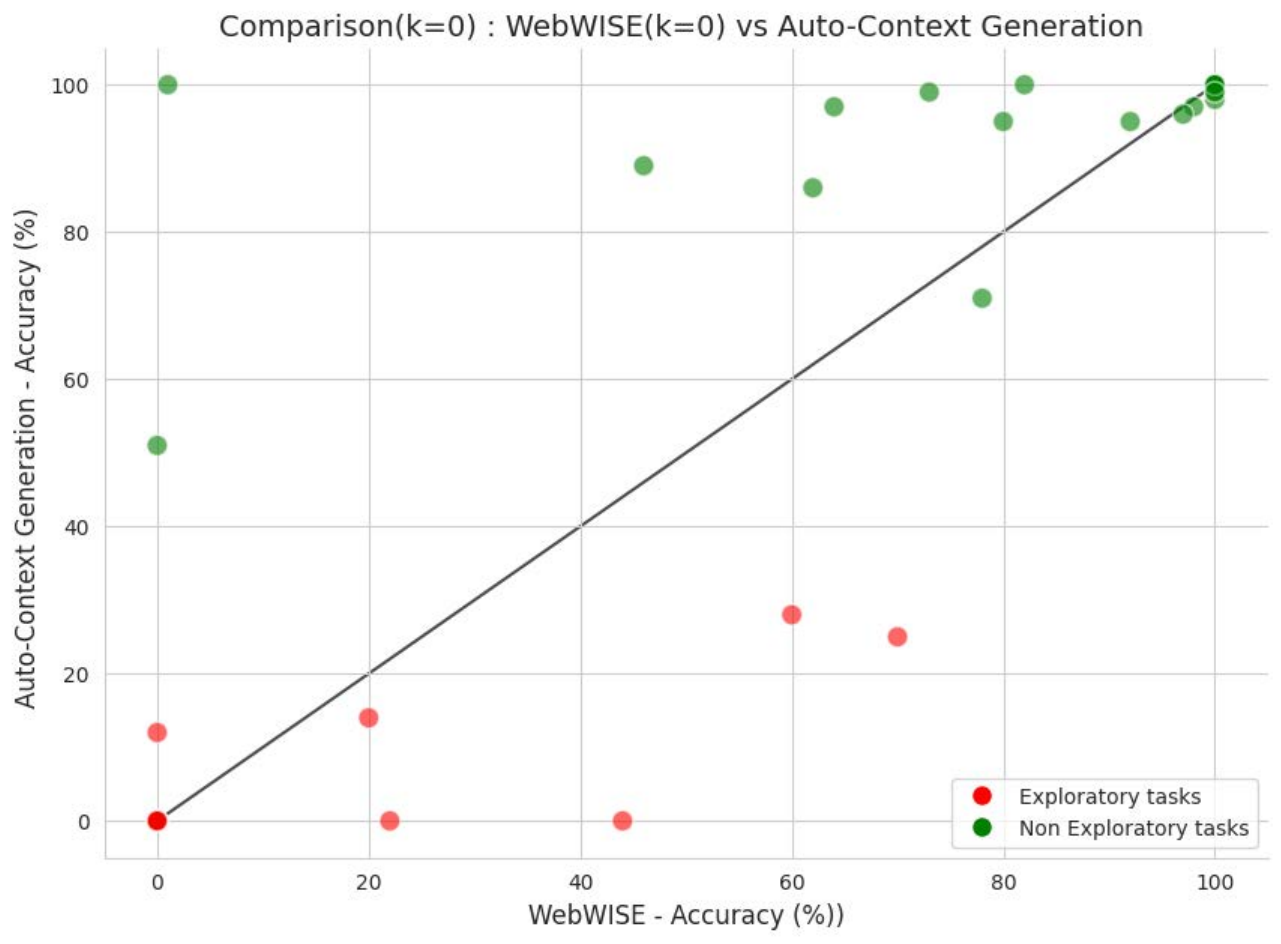}
         \caption{Comparison of WebWISE (x-axis) vs. Auto-Context Generation (ACG) (y-axis) at k=0 across various tasks. Orange dots indicate tasks requiring exploration, such as those where the target button isn't present initially in the DOM elements, whereas green dots represent non-exploratory tasks. A noticeable trend is that Auto-Context Generation's (ACG's) performance diminishes on exploratory tasks. When ACG finds a correct solution during the zero-shot trials, it will continue to use the same solution found instead of generalizing from it.}
         \label{fig:sc4}
     \end{subfigure}
     \caption{Addtional Scatter plots comparing performance across all tasks for different methods. }
\end{figure}

We display additional scatter plots in Figures \ref{fig:sc1}, \ref{fig:sc2}, \ref{fig:sc3}, and \ref{fig:sc4}.





\subsection{Task Analysis}
\begin{figure}[ht]
     \centering
     \begin{subfigure}[b]{0.45\textwidth}
         \centering
         \includegraphics[width=\textwidth]{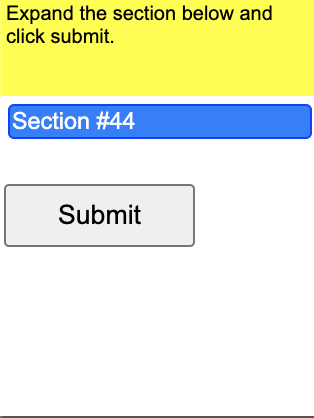}
         \caption{The target is visible for this task. Initial DOM elements contain all relevant information.}
         \label{target_visible}
     \end{subfigure}
     \hfill
     \begin{subfigure}[b]{0.45\textwidth}
         \centering
         \includegraphics[width=\textwidth]{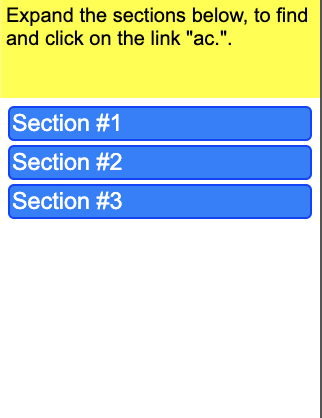}
         \caption{To get the relevant information, different sections have to be expanded. Each time a section is expanded, the DOM elements are changed.}
         \label{target_not_visible}
     \end{subfigure}
\caption{Comparison of tasks where target button is visible and not visible in the initial set of DOM elements.}
\label{fig:target_visible}
\end{figure}

\begin{figure}[ht]
     \centering
     \begin{subfigure}[b]{0.45\textwidth}
         \centering
         \includegraphics[width=\textwidth]{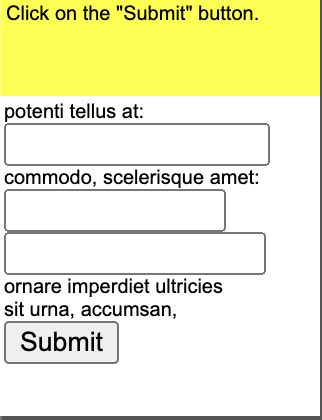}
         \caption{An example of where incorrect answers are \textbf{NOT} present. There is only a single button to click.}
         \label{one_button}
     \end{subfigure}
     \hfill
     \begin{subfigure}[b]{0.45\textwidth}
         \centering
         \includegraphics[width=\textwidth]{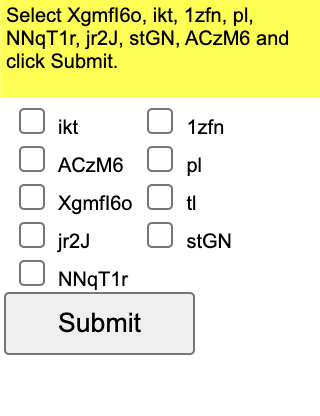}
         \caption{An example where incorrect answers are present. Eg. only some of the checkboxes should be clicked on.}
         \label{target_not_visible1}
     \end{subfigure}
\caption{Comparison of tasks with and without incorrect answers.}
\label{fig:incorrect}
\end{figure}

In the results section, we categorized the tasks based on the number of pre-defined functions needed. However, we also introduce two alternative methods for classifying the tasks. We provide a comprehensive table, Table \ref{tab:task_class}, that presents all the tasks along with their respective classifications.
\paragraph{Incorrect Answers Present} 
One way we can classify the tasks is, if there are incorrect answers present. Tasks that include incorrect answers are characterized by the presence of multiple clickable buttons, as opposed to tasks with a single button. An illustration showcasing a task with and without incorrect answers is provided in Figure \ref{fig:incorrect}.
\paragraph{Target Button not in Initial DOM Elements}
A second way we can classify tasks is based on the initial set of DOM elements. In simpler tasks, the initial set of DOM elements already provides all the necessary information to perform the task. However, for more complex multi-step tasks, it is required to perform at least one additional step to access the DOM elements that contain the target button or the information needed to execute the task correctly. Typically, this additional step involves clicking on a button that triggers a screen change or reveals the relevant elements. Figure \ref{fig:target_visible} provides an example illustrating this concept.

\begin{table}[!ht]
\small 
\centering
\resizebox{\textwidth}{!}{
\begin{tabularx}{\textwidth}{*{4}{X}}
Tasks                       & Number of Functions& Incorrect Answers Visible (Y/N) & Target Button Not in Initial DOM (Y/N) \\
\toprule 
click-button-sequence       & 1          & N  &  N   \\
click-button                & 1          &  N & \textbf{Y}     \\
click-checkboxes-large      & Variable          & \textbf{Y} &  N   \\
click-checkboxes-soft       & Variable        & \textbf{Y} & N   \\
click-checkboxes-transfer   & Variable      & \textbf{Y} &     N \\
click-checkbox              & Variable         & \textbf{Y} &   N   \\
click-collapsible-2         & Variable          & \textbf{Y} & \textbf{Y}     \\
click-collapsible           & 2          & N &  N   \\
click-dialog-2              & 1        & \textbf{Y} &  N   \\
click-dialog                & 1          & N &    N \\
click-link                  & 1          & N & N \\
click-option                & 2         & \textbf{Y} &  N    \\
click-tab-2-hard            & Variable    &\textbf{Y}  & \textbf{Y}    \\
click-tab2-easy             & 2          & \textbf{Y}  &  N \\
click-tab2-medium           & 2          & \textbf{Y}  &  N \\
click-tab-2                 & Variable          & \textbf{Y} & \textbf{Y}    \\
click-tab                   & 1          & N &  N    \\
click-test-transfer         & 1          & \textbf{Y}  & N  \\
click-test-2                & 1          & \textbf{Y} &  N \\
click-test                  & 1         & N  & N    \\
click-widget                & 1         & \textbf{Y} &   N \\
enter-date                  & 3         & N & N     \\
enter-password              & 3         & N & N   \\
enter-text-dynamic          & 3          & N & N \\
enter-text-2                & 3          &  N & N  \\
enter-text                  & 3          & N & N  \\
enter-time                  & 2         & N & N    \\
focus-text-2                & 3         & \textbf{Y} & N   \\
focus-text                  & 3          & N & N    \\
guess-number                & Variable          & N & N  \\
login-user                  & 3         & N & N \\
multi-layouts               & 1         & N & N   \\
use-autocomplete            & 1          & N & N  \\
grid-coordinate             & 1          & N & N     \\
simple-arithmetic           & 2         & N & N   \\
simple-algebra              & 2          & N & N    \\
navigate-tree               & Variable          & \textbf{Y} & \textbf{Y}    \\
search-engine               & Variable         & \textbf{Y} & \textbf{Y}  \\
find-word                   & Variable         & \textbf{Y} & N   \\
email-inbox-forward-nl-turk & 3         & \textbf{Y} & \textbf{Y}     \\
email-inbox-forward-nl      & 3        & \textbf{Y} & \textbf{Y}   \\
email-inbox-nl-turk         & Variable          & \textbf{Y} & \textbf{Y}  \\
email-inbox                 & Variable         & \textbf{Y} & \textbf{Y}    \\
terminal                    & Variable         & \textbf{Y} & \textbf{Y}  \\
click-pie                   & 2       & \textbf{Y}  & \textbf{Y}   \\
read-table                  & 2         & N  & N   \\
read-table-2                & 2         & N  & N \\
text-transform              & 1         & N  & N\\  
\end{tabularx}}
\caption{Classification of all tasks. Y is yes and N is no.}
\label{tab:task_class}
\end{table}

\subsubsection{Task Failures}
In this section, we delve into the reasons behind the failure of certain tasks. It is worth noting that some failures are attributed to the different task classifications we discussed earlier.

One specific example is the 'click-dialog' task, as demonstrated in Table 3. When using our WebWISE method with only one example, we achieve perfect accuracy. However, for the 'click$-$dialog$-$2' task, the accuracy drops to $76\%$. This discrepancy can be attributed to a particular sub-task where the task description instructs the user to click on the 'x' symbol to close a dialog. Unfortunately, the language model may not fully comprehend that the symbol 'x' typically represents the close function. As a result, this lack of understanding leads to failures in executing this specific sub-task.

\paragraph{Email Tasks}
Among the tasks we evaluated, some simulate performing actions in an email mobile app. Notably, the performance varies across these tasks, particularly those that begin with the word "email." Tasks starting with "email-inbox-forward" tend to have better performance compared to tasks without the word "forward". This variation can be explained by the scope of the individual tasks. In tasks with "forward" in the prompt, there is only one specific action that needs to be performed: forwarding an email. However, in tasks without the word "forward," there can be one of four possible actions: forwarding, starring (marking as important), deleting, and replying to an email. To maintain consistency with the other tasks, we used a single in-context example with only one of the actions for the email-related tasks. Consequently, our method can only successfully execute the action mentioned in the prompt. This limitation explains the discrepancy in performance, as tasks without the word "forward" require the model to perform one out of multiple potential email actions, while tasks with "forward" solely focus on forwarding an email.

\paragraph{Terminal}
One common source of failure in tasks involving a terminal is the model's lack of knowledge on how to execute a command by pressing the enter key. The task only succeeds when we provide an example because we use the action $enter\_text\_action("CommandToBeEntered\backslash n", observation)$, which simulates pressing the enter key after entering the command. To address this issue, a potential solution in the zero-shot scenario is to develop primitive functions specifically for key actions such as "ENTER," "BACKSPACE," or "DELETE." By incorporating these primitive functions into the model, it can learn to perform key-related actions more effectively in terminal-based tasks. 

\paragraph{Search Engine and Text Transform}
Failures in tasks like "Search-Engine" and "Text-transform" can be attributed to the limitations of the $getSummary()$ function in capturing the order of DOM elements. For example, in the search engine task, the instruction may involve clicking on the 8th search result on a webpage. While the DOM elements may accurately capture the order of search results, the filtered DOM elements fail to preserve this order, leading to task failures. A similar observation can be made for tasks like "Text Transform." To address this issue, it becomes necessary to develop an adaptive $getSummary()$ function that can extract the most relevant elements while also preserving their order within the DOM. Alternatively, approaches involving the use of image input modalities in the form of RGB vectors could be explored to overcome these limitations. Such approaches can provide a visual representation of the webpage, enabling the model to better understand the layout and order of the elements present.

\subsection{Additional Task Analysis}

\begin{figure}[!ht]
  \centering
  \includegraphics[scale=0.45]{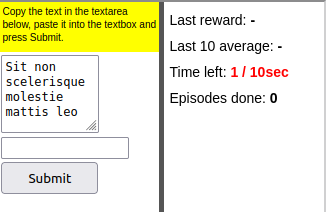}
  \caption{An example of the copy-paste task}
  \label{fig:copypaste1}
\end{figure}

\begin{figure}[!ht]
  \centering
  \includegraphics[scale=0.45]{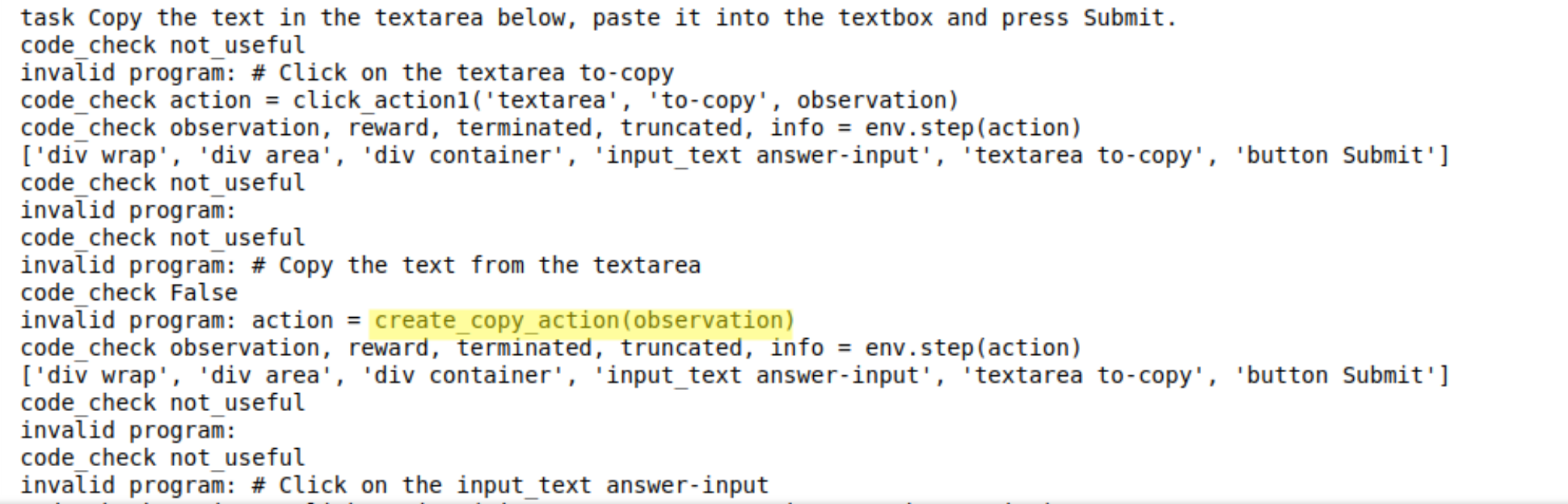}
  \caption{Generated program for the copy-paste task in Zero Shot scenario}
  \label{fig:copypaste2}
\end{figure}

\begin{figure}[!ht]
  \centering
  \includegraphics[scale=0.45]{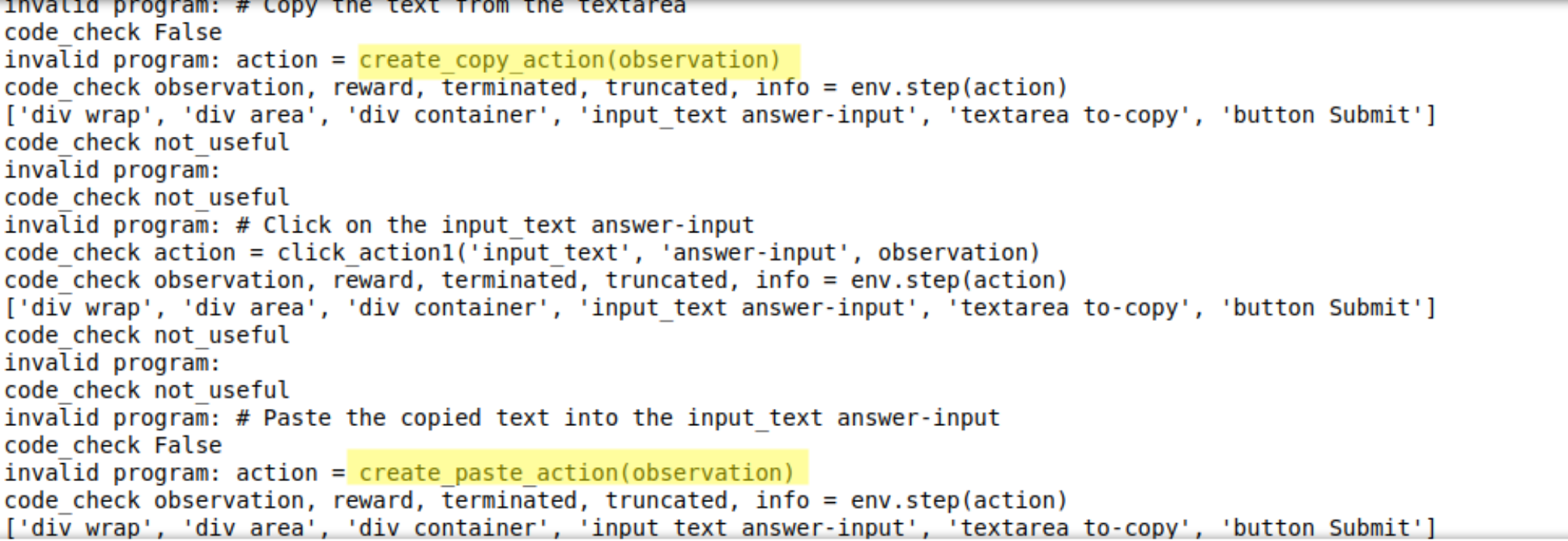}
  \caption{Generated program for the copy-paste task in Zero Shot scenario}
  \label{fig:copypaste3}
\end{figure}

In tasks like "Copy-Paste," the objective is to copy text from the task interface and paste it into an empty text field, as illustrated in Figure \ref{fig:copypaste1}. However, we have not implemented the copy-paste function for the language model (LLM) to interact with the environment and complete the task. We noticed that the LLM makes references to functions like "$create\_copy\_action$" and "$create\_paste\_action$," which, if implemented, could have facilitated the solution (as shown in Figures \ref{fig:copypaste2},\ref{fig:copypaste3}). Some additional functions such as clicking on specific coordinates can also be potentially implemented to improve the LLM's capabilities in handling tasks like these. 

\subsection{Results Comparison with RCI paper}
In Table \ref{tab:rci}, we compare our results with the concurrent work from RCI paper. 
\begin{table}[!ht]
\small 
\centering
\resizebox{\textwidth}{!}{
\begin{tabularx}{\textwidth}{c*{3}{X}}
Tasks                       & WebWISE (k=1) & RCI  & RCI k \\
\toprule 
choose-list                 & n/a           & \textbf{1.00} & 0     \\
click-button-sequence       & \textbf{1.00} & \textbf{1.00} & 2     \\
click-button                & \textbf{1.00}          & \textbf{1.00} & 1     \\
click-checkboxes-large      & \textbf{1.00}          & 0.94 & 1     \\
click-checkboxes-soft       & \textbf{0.78}          & 0.72 & 1     \\
click-checkboxes-transfer   & \textbf{1.00}          & \textbf{1.00} & 2     \\
click-checkbox              & \textbf{1.00}          & \textbf{1.00} & 2     \\
click-collapsible-2         & \textbf{0.66}          & 0.62 & 2     \\
click-collapsible           & \textbf{1.00}          & \textbf{1.00} & 1     \\
click-color                 & n/a           & \textbf{1.00} & n/a   \\
click-dialog-2              & 0.74          & \textbf{1.00} & 3     \\
click-dialog                & \textbf{1.00}          & \textbf{1.00} & 1     \\
click-link                  & \textbf{1.00}          & \textbf{1.00} & n/a   \\
click-menu                  & n/a           & \textbf{1.00} & 1     \\
click-option                & \textbf{1.00}          & \textbf{1.00} & 1     \\
click-scroll-list           & n/a           & \textbf{1.00} & 1     \\
click-shades                & n/a           & \textbf{1.00} & 1     \\
click-shape                 & n/a           & \textbf{0.98} & 2     \\
click-tab-2-hard            & 0.68          & \textbf{0.76} & 6     \\
click-tab2-easy             & \textbf{0.96}          & n/a  & n/a   \\
click-tab2-medium           & \textbf{0.52}          & n/a  & n/a   \\
click-tab-2                 & \textbf{0.78}          & 0.74 & 1     \\
click-tab                   & \textbf{1.00}          & \textbf{1.00} & 1     \\
click-test-transfer         & \textbf{1.00}          & n/a  & n/a   \\
click-test-2                & \textbf{1.00}          & \textbf{1.00} & n/a   \\
click-test                  & \textbf{1.00}          & \textbf{1.00} & 1     \\
click-widget                & \textbf{0.98}          & \textbf{0.98} & 3     \\
count-shape                 & n/a           & \textbf{0.40} & n/a   \\
enter-date                  & \textbf{1.00}          & 0.96 & 4     \\
enter-password              & \textbf{1.00}          & \textbf{1.00} & n/a   \\
enter-text-dynamic          & \textbf{1.00}          & \textbf{1.00} & n/a   \\
enter-text-2                & \textbf{1.00}          & n/a  & n/s   \\
enter-text                  & \textbf{1.00}          & \textbf{1.00} & n/s   \\
enter-time                  & \textbf{1.00}          & \textbf{1.00} & 2     \\
focus-text-2                & \textbf{1.00}          & \textbf{1.00} & n/a   \\
focus-text                  & \textbf{1.00}          & \textbf{1.00} & 1     \\
guess-number                & \textbf{0.84}          & 0.20 & n/a   \\
identify-shape              & n/a           & \textbf{0.76} & n/a   \\
login-user-popup            & n/a           & \textbf{0.68} & n/a   \\
login-user                  & \textbf{1.00}          & \textbf{1.00} & n/a   \\
multi-layouts               & \textbf{0.78}          & 0.72 & n/a   \\
multi-orderings             & n/a           & \textbf{1.00} & n/a   \\
social-media-all            & n/a           & \textbf{1.00} & 4     \\
social-media-some           & n/a           & \textbf{0.90} & n/a   \\
social-media                & n/a           & \textbf{0.98} & 2     \\
tic-tac-toe                 & n/a           & \textbf{0.56} & n/a   \\
use-autocomplete            & 0.92          & \textbf{0.58} & n/a   \\
use-spinner                 & n/a           & \textbf{0.88} & 8     \\
grid-coordinate             & \textbf{1.00}          & \textbf{1.00} & 1     \\
simple-arithmetic           & \textbf{1.00}          & n/a  & n/a   \\
simple-algebra              & \textbf{1.00}          & \textbf{1.00} & 1     \\
navigate-tree               & \textbf{0.92}          & 0.86 & 1     \\
search-engine               & 0.20          & \textbf{1.00} & 22    \\
find-word                   & 0.00          & n/a  & n/a   \\
email-inbox-forward-nl-turk & 0.92          & \textbf{0.94} & 3     \\
email-inbox-forward-nl      & 0.96          & \textbf{1.00} & 3     \\
email-inbox-nl-turk         & 0.18          & \textbf{0.98} & 4     \\
email-inbox                 & 0.21          & \textbf{0.98} & 6     \\
email-inbox full example    & n/a           & n/a  & n/a   \\
terminal                    & 0.93          & \textbf{1.00} & 1     \\
click-pie                   & \textbf{0.80}          & n/a  & n/a   \\
read-table                  & \textbf{0.86}          & n/a  & n/a   \\
read-table-2                & \textbf{0.82}          & n/a  & n/a   \\
text-transform              & \textbf{1.00}          & n/a  & n/a  
\end{tabularx}}
\caption{Comparison of WebWISE and RCI. The last column lists the k value for RCI if available.}
\label{tab:rci}
\end{table}

\subsection{Sensitivity to prompts}

In our analysis, we have observed that the language model (LLM) exhibits sensitivity to the prompts provided, including the system message and task message. To delve deeper into this phenomenon, we conducted a case study where we deliberately varied the task message and examined the resulting performance of the generated programs.

By systematically altering the task message, we aimed to understand how the LLM's responses and generated programs are influenced by different prompts. This sensitivity to prompts highlights the importance of carefully crafting and designing prompts to elicit the desired behavior and improve the overall performance of the language model.

\paragraph{Task Message}

\begin{lstlisting}[language=Python,basicstyle=\tiny, caption=Experiments on different task messages for API,label={lst:api4},belowskip=-0.8 \baselineskip]

task_message_1="""This task is a multi-step challenge. To successfully complete it, you need to be aware of the current state of the environment and the user input. Before performing any action, carefully observe and analyze the environment to determine whether further actions are required. When exploring and trying different actions, ensure that you select appropriate actions and arguments for the functions based on the current environment. Focus on efficiently reaching a solution by checking if the task can be solved with the current user input and environment state before taking any further steps, and by using correct actions and arguments for each function."""

task_message_2="""Next task is a multi-step task, directly performing a series of actions may not solve the task. Need to observe the changes in the user input before and after performing any action to see if 
further actions need to be made to solve the task or not"""

task_message_3= """Next task is a multi-step task, directly performing a series of actions may not solve the task. Need to observe the changes in the user input before and after performing any action to see if 
further actions need to be made to solve the task or not. So, explore and try different actions and figure out a way to solve the task, but at every step check if you are able to solve the task with the current user input before taking the action."""

task_message_4="""Your next task is a multi-step challenge. To successfully complete it, carefully observe and analyze the changes in user input before and after performing any action. This will help determine whether further actions are necessary. While it's important to explore and try various actions, always assess whether the task can be solved with the current user input before taking additional steps. Focus on efficiently reaching a solution without excessive exploration when a satisfactory outcome is already achievable. 
"""

task_message_5="""The upcoming task is a multi-step challenge that requires you to pay close attention to the current user input. Your goal is to efficiently reach a solution by performing appropriate actions based on the present situation. Before taking any action, analyze the user input to determine if further actions are necessary. Explore and try the next action, but always ensure they are necessary, relevant to the current state and have the correct arguments for the functions. Continually assess the situation to check if the task can be solved with the current user input and environment state before proceeding further."""

task_message_6="""The upcoming task is a multi-step challenge. To successfully complete it, you must be aware of the current state of the environment and the user input. Before performing any action, carefully observe and analyze the environment to determine whether further actions are required. When selecting actions, ensure that you only perform actions if the current user input has the necessary elements. Focus on efficiently reaching a solution by trying to solve the task with the current user input and environment state before considering further exploration. Only explore and try different actions if the task cannot be solved with the current state. Make sure to use correct actions and arguments for each function based on the current environment."""

task_message_7="""The this task is a multi-step challenge. To successfully complete it, you must be aware of the current state of the environment and the objects in the image. Before performing any action, carefully observe and analyze the environment to determine whether further actions are required. When selecting actions, ensure that you only perform actions if the objects in the image have the necessary elements. Focus on efficiently reaching a solution by trying to solve the task with the current objects in the image and environment state before considering further exploration. Only explore and try different actions if the task cannot be solved with the current state. Make sure to use correct actions and arguments for each function based on the current environment."""

task_message_8="""The upcoming task is a multi-step challenge. Observe and analyze the environment and objects in the image before performing any action. Select actions based on the current state and ensure they are relevant to the objects in the image. Focus on solving the task with the current state, and only explore further if necessary. Use correct actions and arguments for each function, and be mindful of the environment during the process."""

task_message_9="""This task is a multi-step challenge. Observe and analyze the user input which contains the objects in the image before performing any action. Select actions based on the current state and ensure they are relevant to the objects in the image. Try other actions if and only if you are not able to solve the task with the current user input. Use correct actions and arguments for each function, and be mindful of the environment(user input) during the process."""

task_message_10="""In this multi-step task, stay aware of the environment and user input. Observe and analyze before acting. As you try actions, choose suitable functions and arguments. Focus on efficiency: check if the task is solvable with current input and environment before proceeding. Converge toward the objective by using correct actions and arguments, and be cautious to avoid divergence."""

task_message_11="""This task involves a multi-step challenge, which can be accomplished by following these succinct steps:

1)Examine the environment by analyzing objects in the image from user input.
2)Determine if the desired element from the task image is present in the current objects.
3)If not, perform necessary actions (e.g., clicking, scrolling) to make the element available.
4)Iterate steps 1-3 until the desired element is found and can be clicked or interacted with.
5)Once the element is available and visible, execute the appropriate action on it."""


task_message_12="""
This task is a multi-step challenge, which can be accomplished by following these steps:
1)You should solve it step by step. 
2)Before performing any action, determine if the desired element from the task is present in the Objects in Image.
3)If and only if the desired object is not there, say the phrase "The desired object is not there" 
4)Then explore and perform other actions (e.g., clicking, scrolling) to see if the desired element is available in other states.
5)Iterate steps 2-4 until the desired element is found and can be clicked or interacted with.
"""

task_message_13="""
This task involves a multi-step challenge, which can be accomplished by following these succinct steps:
1)Determine if the desired element from the task image is present in the Object  in Image.
2)If its not, explore and perform other actions (e.g., clicking, scrolling) to see if the element is available in other states.
3)Iterate steps 1-2 until the desired element is found and can be clicked or interacted with.
4)Once the element is available and visible, execute the appropriate action on it.
"""


\end{lstlisting}

\begin{table}[H]
\begin{tabularx}{\textwidth}{l*{3}{X}}
\textbf{Various Task Message} & \textbf{Average performance on WebWise k=0} & {\textbf{Average performance on WebWise k=1}} \\
\toprule 
Task Message 1                & .62                                                            & .85                                                            \\
Task Message 2                & .52                                                         & .72                                                            \\
Task Message 3                & .50                                                            & .76                                                           \\
Task Message 4                & .54                                                            & .74                                                           \\
Task Message 5                & .48                                                            & .74                                                          \\
Task Message 6                & .54                                                         & .76                                                          \\
Task Message 7                & .56                                                           & .76                                                            \\
Task Message 8                & .56                                                         & .78                                                           \\
Task Message 9                & .58                                                            & .78                                                          \\
Task Message 10               & .60                                                        & .80                                                           \\
Task Message 11               & .56                                                            & .76                                                          \\
Task Message 12               & .54                                                            & .78                                                           \\
Task Message 13               & .58                                                            & .80                                                           
\end{tabularx}
\caption{WebWISE k=0 and k=1 performance for the different task messages }
\end{table}

The impact of variation in the task message on the performance of the proposed approach, WebWISE, is evaluated here. Although the overall meaning of the "task message" remains consistent, minor differences in sentence structure and syntax can affect the model's performance, particularly in certain tasks requiring multiple steps. While the average performance across tasks may appear similar, there are significant variations in accuracy among individual multistep tasks, with some showing a high standard deviation of 62\%. Thus we could develop methods that can handle prompt variation robustly.

Furthermore, a deeper analysis of specific tasks like "Simple-Algebra" and "Navigate-Tree" revealed that providing the complete DOM elements as input improves performance. This improvement is attributed to the presence of keywords like "math-question" and "folder" within the DOM elements. Although these keywords may not be essential for executing actions within the web interface, they play a crucial role in providing contextual information to the language model. Consequently, the model gains a better understanding of the broader task it needs to accomplish at any given moment.

\end{document}